\documentclass{article}
\usepackage[preprint]{corl_2026} 
\usepackage{graphicx}
\usepackage{float}
\usepackage{booktabs}
\usepackage{enumitem}
\usepackage{diagbox}
\usepackage{wrapfig}
\usepackage{pifont}
\usepackage{amsmath}
\usepackage{amssymb}
\usepackage[T1]{fontenc}       
\usepackage{tikz}
\usepackage{caption}
\usepackage{booktabs}
\usepackage{multirow}
\usepackage{svg}
\usepackage{algorithm}
\usepackage{algpseudocode}

\usepackage{amsthm}
\newtheorem{proposition}{Proposition}[section]

\AtBeginDocument{\nolinenumbers}

\title{ATHENA: Accelerated Multi-Task Heterogeneous Influence Functions for Robot Data Curation}

\author{
  \begin{minipage}{\textwidth}
  \centering\bfseries
  Tao Xu$^{1,2}$,
  Jiaxin Wang$^{2,3}$,
  Runhao Zhang$^{2}$,
  Jiayi Guan$^{1}$,
  Xianchao Zeng$^{2}$, \\[0.1em]
  Weixi Song$^{2}$,
  Xinyu Zhou$^{2}$,
  Zhetao Chen$^{2}$,
  Guang Chen$^{1,2}$,
  Yong-Lu Li$^{2,4,\dagger}$
  \end{minipage}\\[1.0em]
\begin{minipage}{\textwidth}
\centering\normalfont\footnotesize
  $^{1}$Tongji University, 
  $^{2}$Shanghai Innovation Institute,
  $^{3}$Xi'an Jiaotong University, \\[0.15em]
  $^{4}$Shanghai Jiao Tong University
  \end{minipage}\\[-2.8em]
}

\makeatletter
\renewcommand{\@notice}{}
\patchcmd{\@maketitle}{\vskip 0.3in \@minus 0.1in}{\vskip 0.12in \@minus 0.05in}{}{}
\makeatother

\begin{document}
\hypersetup{pdfauthor={Tao Xu, Jiaxin Wang, Runhao Zhang, Jiayi Guan, Xianchao Zeng, Weixi Song, Xinyu Zhou, Zhetao Chen, Guang Chen, Yong-Lu Li}}

\maketitle

\begin{tikzpicture}[remember picture,overlay]
  \node[anchor=south west, inner sep=0] 
    at ([xshift=\dimexpr1in+\oddsidemargin\relax,yshift=0.55in]current page.south west) {%
    \parbox{\textwidth}{\footnotesize\raggedright
    $^{\dagger}$Corresponding author: \texttt{yonglu.li@sjtu.edu.cn}}%
  };
\end{tikzpicture}

\begin{abstract}
    In robot imitation learning, influence functions provide a principled approach to quantify each demonstration's effect on robot task outcomes, yet scaling them to billion-parameter Vision-Language-Action (VLA) models is limited by computational and multitask bottlenecks. To this end, we propose ATHENA, an influence function framework tailored for multitask VLA data curation at a billion-parameter scale. Concretely, it leverages the Kronecker structure of linear-layer gradients to reduce projection cost, and approximates dense Hessian inversion with a rank-$r$ Random Truncated Approximation, achieving about a 313.4$\times$ speedup in influence computation. Furthermore, ATHENA formulates global and local interactive influence to balance data curation across 50 jointly trained tasks. Extensive evaluations on RoboTwin 2.0 and real-robot deployment, covering 9.34 and 6.90 hours of demonstrations, respectively, show that ATHENA matches or exceeds full-data joint fine-tuning using only 50\% of demonstrations in simulation and 66.7\% of data across six real-robot tasks. Overall, ATHENA demonstrates its effectiveness for data curation in billion-parameter multitask VLA fine-tuning. Project website: \href{https://sii-quantum.github.io/ATHENA.github.io/}{this URL}.

\end{abstract}
\section{Introduction}

Vision-Language-Action (VLA) models have shown promising results in robotic manipulation from large-scale robot demonstrations~\cite{zitkovich2023rt2,kim2024openvla,black2024pi0}, but their performance also depends critically on data quality~\cite{lin2025datascaling,NEURIPS2023_fe692980,mu2024robotwin}. Naively scaling demonstration data for fine-tuning can incur high costs while yielding limited or even degraded performance~\cite{lee2026quality_cupid2, agia2025cupid}. This motivates a central question for VLA fine-tuning: \emph{which demonstrations should be retained to maximize the performance of VLA models?}

Answering this question requires efficient, data-centric methods to identify high-quality subsets of fine-tuning demonstrations.
However, existing approaches face challenges when applied to billion-parameter VLA models. Specifically, traditional leave-one-out data valuation requires retraining the models for each candidate subset, which incurs prohibitive computational cost~\cite{ghorbani2019data}.
Expert-based or data distillation curation methods are more efficient, but they often do not capture the causal effect of training data on downstream policy performance~\cite{NEURIPS2023_fe692980,hejna2025robotdatacurationmutual,mandlekar2021robomimic}.
In contrast, gradient-based influence functions~\cite{koh2017understanding} provide a principled and interpretable framework without repeated retraining. For example, CUPID~\cite{agia2025cupid} applies influence functions to estimate the effect of each robot demonstration, curating high-quality subsets that match or exceed full-data training. 

However, CUPID is limited to small imitation policies (24M parameters) and single-task curation~\cite{agia2025cupid,chi2023diffusion}, preventing direct scaling to billion-parameter multitask VLA models such as $\pi_0$~\cite{black2024pi0}. Specifically, per-sample gradient computation incurs $\mathcal{O}(DP)$ cost, where $D$ is the number of model parameters and $P$ is the projection dimension, and dense Hessian inversion entails $\mathcal{O}(NP^2 + P^3)$ complexity, where $N$ is the number of training samples, which together limit the scalability of influence functions~\cite{park2023trak,choe2026logra,mlodozeniec2025influenceorl}. Moreover, because influence scores are computed greedily~\cite{NEURIPS2019_a78482ce,pmlr-v267-feng25l}, naively extending single-task attribution to multitask settings may produce imbalanced curation across tasks~\cite{tu2025measuring,dass2026datamil}.

To overcome these challenges, we introduce ATHENA, an influence function framework for multitask VLA data curation at a billion-parameter scale.
Concretely, it improves computational efficiency along two axes: gradient projection at each layer and Hessian approximation. For gradient projection, ATHENA leverages the Kronecker structure of linear layer gradients to perform low-rank projections at each layer, achieving a square root reduction in projection cost at each layer, from $\mathcal{O}(DP)$ to $\mathcal{O}(\sqrt{DP})$. For Hessian approximation, ATHENA approximates dense Hessian inversion with a rank-$r$ Random Truncated Approximation, lowering the leading cost from $\mathcal{O}(N\cdot P^2 + P^3)$ to $\mathcal{O}(N\cdot P\cdot r)$. This reduces influence computation from 8054.6 to 25.7 GPU-hours, corresponding to about a 313.4$\times$ speedup. Furthermore, ATHENA formulates multitask influence interaction with global and local interactive influence measures for balanced curation over 50 jointly trained tasks, improving the return on investment (ROI) of data curation. 

Our contributions are summarized as follows: (1) We propose ATHENA, an influence function framework for multitask VLA data curation at billion-parameter scale. It leverages Kronecker-structured gradient projection and a rank-$r$ Random Truncated Approximation to significantly reduce the cost of influence function computation, achieving a $313.4\times$ speedup. (2) We present a global and local influence design that balances multitask data curation and improves ROI across 50 jointly trained tasks. (3) We evaluate ATHENA in RoboTwin 2.0 simulation and real-robot deployment, where it matches or exceeds full-data joint fine-tuning using only $50\%$ of demonstrations in simulation and $66.7\%$ of data across six real-robot tasks. These results demonstrate its effectiveness for VLA fine-tuning data curation.

\section{Related Work}
\paragraph{Robot Imitation Learning.}
Imitation learning has become central to robot research~\cite{zeng21aTransporterNetworks,journals/corr/abs-2109-00137,pmlr-v164-jang22a,khazatsky2025droidlargescaleinthewildrobot,ICRA2024OpenXEmbodiment}. Early methods such as ACT~\cite{zhao2023learning} and Diffusion Policy~\cite{chi2023diffusion} achieve excellent performance under task-specific training, but training one policy per task reduces the ROI of joint robot learning and limits multitask data scaling. In contrast, VLA models leverage large-scale pretrained knowledge and multitask fine-tuning to improve both performance and ROI~\cite{fan2025longvla,wen2025dexvla}. However, their performance depends not only on data scale, but also on demonstration quality~\cite{pmlr-v229-kuhar23a,pmlr-v205-gandhi23a}. Large-scale yet redundant datasets may even harm VLA performance~\cite{yang2026less,yu2026frameskiplearningfewerinformative}, making data curation essential.

\textbf{Robot Data Curation.}
Recent robot data curation methods aim to improve policy performance by identifying high-utility demonstrations~\cite{mandlekar2023mimicgen,garrett2024skillmimicgen,yu2023scalingrobotlearningsemantically}. Leave-one-out valuation and Data Shapley directly estimate data utility but require repeated retraining~\cite{ghorbani2019data}, which incurs unacceptable computational cost~\cite{jia2020efficienttaskspecificdatavaluation}. More efficient methods, including dataset distillation~\cite{wang2020datasetdistillation,DBLP:journals/corr/abs-2006-05929}, quality scoring~\cite{hejna2025robotdatacurationmutual,mandlekar2021robomimic}, retrieval~\cite{du2023behaviorretrievalfewshotimitation,nasiriany2022learningretrievalpriordata,lin2024flowretrieval,ICLR2025_a06e129e}, and mixture learning~\cite{hejna2024remix,sima2026promix}, avoid retraining but rely on proxy objectives rather than the  causal effects on downstream policy performance. Gradient-based curation methods, including CUPID~\cite{agia2025cupid}, QoQ~\cite{lee2026quality_cupid2}, and DataMIL~\cite{dass2026datamil}, estimate demonstration utility for imitation learning, yet scaling them to billion-parameter multitask VLA fine-tuning remains challenging~\cite{yang2026less,pmlr-v267-feng25l}. In practice, as robot imitation learning shifts from small task-specific policies to billion-parameter multitask VLA models, robot data curation also requires an efficient method tailored to VLA fine-tuning.

\label{sec:related}

\section{Preliminaries and Problem Formulation}
\label{sec:prelim}
\subsection{Robot Data Curation with Influence Functions}
\label{sec:if_curation}

We consider a VLA model $\pi_\theta$, $\theta\in\mathbb{R}^D$, fine-tuned on demonstrations $\mathcal{D}=\{\xi_i\}_{i=1}^n$. Each demonstration trajectory consists of sequential state-action pairs, formulated as individual training samples $z_t^i = (s_t^i,a_t^i)$. Thus, a demonstration is denoted as $\xi_i=\{z_t^i\}_{t=1}^{H_i}$, and the total number of training timesteps is $N=\sum_i H_i$. Let $\theta^\star$ denote the fine-tuned parameters and $\mathcal{L}$ the imitation loss. Classical influence functions~\cite{koh2017understanding} estimate the first-order effect of upweighting a single training sample $z$ on a test quantity $f(\hat z;\theta)$, with $H_\theta=\nabla_\theta^2\mathcal{L}(z_t^i;\theta^\star)$:
\begin{equation}
    \Psi_{\mathrm{inf}}(\hat z, z)
    =
    -\nabla_\theta f(\hat z;\theta^\star)^\top
    H_\theta^{-1}
    \nabla_\theta \mathcal{L}(z;\theta^\star).
    \label{eq:if_classic}
\end{equation}

However, robot demonstrations are sequential trajectories whose influence scores should reflect closed-loop task performance rather than step-wise losses~\cite{agia2025cupid,lee2026quality_cupid2}. Accordingly, motivated by CUPID's~\cite{agia2025cupid} closed-loop attribution principle, we instantiate the action-level influence $\Psi_{a\text{-inf}}(\hat{z}, z)$ for the flow-matching VLA model $\pi_0$, and aggregate it into demonstration-level performance influence:
\begin{equation}
    \widehat{\Psi}_{\pi\text{-inf}}(\xi_i)
    =
    \frac{1}{m}
    \sum_{\tau_j\in\mathcal{E}}
    \frac{R(\tau_j)}{H_i}
    \sum_{\hat{z}\in\tau_j}
    \sum_{z\in\xi_i}
    \Psi_{a\text{-inf}}(\hat{z}, z),
    \label{eq:perf_inf}
\end{equation}
where $\mathcal{E}$ contains $m$ evaluation rollouts, $R(\tau_j)\in\{1,-1\}$ is the return of rollout $\tau_j$, $H_i$ is the length of demonstration $\xi_i$, and $\Psi_{a\text{-inf}}$ is mathematically derived in Appendix~\ref{app:action_influence}.

\subsection{Scalability Barriers for Billion-Parameter VLA Curation}
\label{sec:barriers}

Eq.~\eqref{eq:perf_inf} defines a closed-loop performance influence function for robot data curation. However, scaling it to billion-parameter multitask VLA models faces two main bottlenecks: costly influence estimation at the model parameter and data scale, and imbalanced greedy selection under heterogeneous task distributions.

\paragraph{Scaling bottleneck.}
Naively computing projected gradients requires materializing
$g_i=\nabla_\theta\ell(z_i;\theta^\star)\in\mathbb{R}^{D}$ and projecting it
into a $P$-dimensional feature space. Let
$\Omega\in\mathbb{R}^{D\times P}$ denote the random projection matrix. The projected gradient feature is:
\begin{equation}
    \phi_i = \Omega^\top g_i .
    \label{eq:projected_grad}
\end{equation}
The vector $\phi_i\in\mathbb{R}^{P}$ denotes the projected feature of $z_i$. This incurs $\mathcal{O}(DP)$ gradient computation and storage cost, which is prohibitive for billion-parameter VLA models~\cite{choe2026logra,li2411delta_logra_trak,zhang2025toward_logra_trak}.

Subsequently, training gradients are stacked into a projected gradient matrix $G\in\mathbb{R}^{N\times P}$, whose $i$-th row is $\phi_i^\top$. Influence scores are computed via a damped Gauss-Newton approximation~\cite{park2023trak}:
\begin{equation}
    \widehat{\psi}(z_{\mathrm{te}},z_{\mathrm{tr}})
    =
    \phi_{\mathrm{te}}^\top
    \left(G^\top G+\lambda I_P\right)^{-1}
    \phi_{\mathrm{tr}}.
    \label{eq:trak_score}
\end{equation}
In Eq.~\eqref{eq:trak_score}, $\phi_{\mathrm{tr}}$ and $\phi_{\mathrm{te}}$ denote the projected gradients of training and evaluation timesteps, $\lambda$ is the damping coefficient, and $I_P$ is the identity matrix. Although this formulation avoids forming the full $D\times D$ Hessian, it still introduces dense computation over the projected training matrix: forming and inverting $G^\top G+\lambda I_P$ costs $\mathcal{O}(NP^2+P^3)$~\cite{tu2025rrinf_haisen, wang2026fast_haisen,li2025did_haisen}, which becomes a severe computational bottleneck as $N$ increases. Therefore, billion-parameter VLA curation requires reducing both projected gradient computation and projected inverse-Hessian approximation costs.

\paragraph{Multitask curation bottleneck.}
For curation over $K$ tasks, estimating influence and fine-tuning independently for each task increases the total computational and storage cost by a factor of $K$, yielding low ROI. A single greedy influence ranking shared across all tasks is more efficient, but tends to favor tasks with stronger gradient signals, leading to imbalanced task coverage~\cite{pmlr-v267-feng25l,tu2025measuring,dass2026datamil}. Multitask VLA curation therefore requires an attribution rule that accounts for both influence across tasks and relevance within each task.

\begin{figure}[t]
    \centering
    \includegraphics[width=\linewidth]{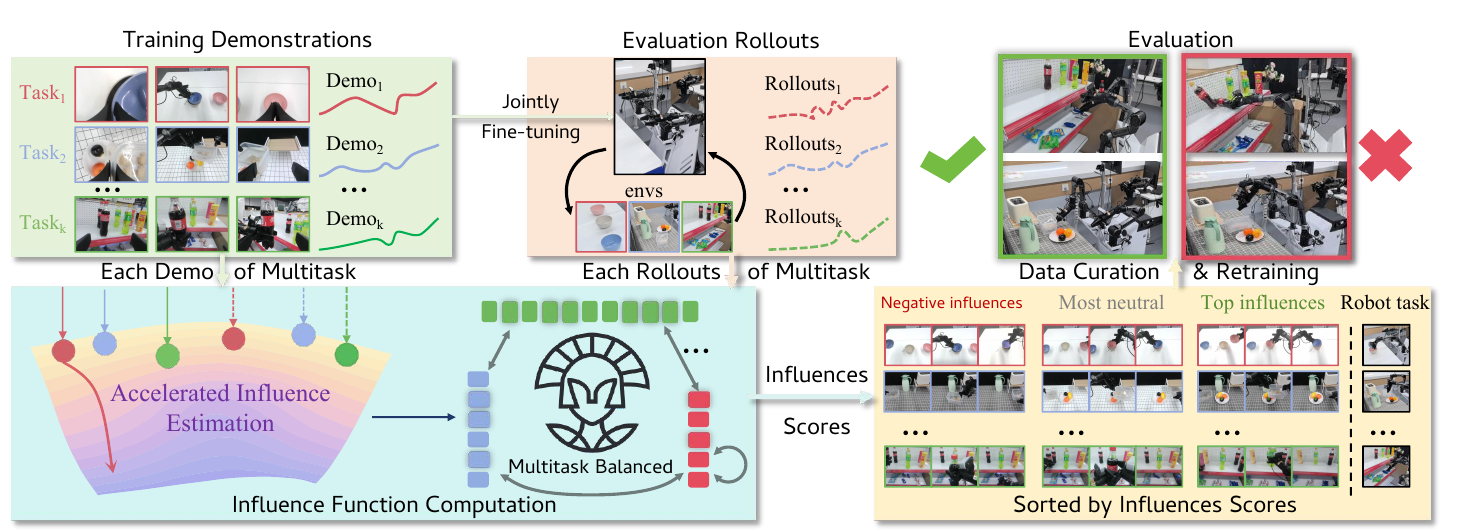}
    \vspace{-0.4cm}
\caption{\textbf{ATHENA pipeline.} Training dataset and closed-loop rollouts from VLA evaluation are fed into an efficient multitask influence computation module to score and rank demonstration importance, guiding high-quality data curation for VLA fine-tuning.
}

    \label{fig:pipeline}
\end{figure}

\section{ATHENA: Scalable Influence Function Data Curation for VLA Models}
\label{sec:athena_method}

To overcome the scalability and multitask bottlenecks in Section~\ref{sec:prelim}, we propose ATHENA, an influence function framework for billion-parameter multitask VLA models. Specifically, ATHENA reduces computational and memory constraints by combining Kronecker compressed per-sample gradient featurization with a Random Truncated Approximation of the inverse Hessian. In addition, ATHENA models both global and local influence interactions, enabling effective multitask curation without greedy task imbalance. Figure~\ref{fig:pipeline} presents an overview of the pipeline.

\subsection{Accelerated Influence Estimation}
\paragraph{Kronecker Compressed Gradient Featurization.}
Following Eq.~\eqref{eq:projected_grad}, let
$g_i=\nabla_\theta\ell(z_i;\theta^\star)\in\mathbb{R}^{D}$ denote the
gradient of the $i$-th training sample. Naive gradient projection computes
$\phi_i=\Omega^\top g_i$ with $\Omega\in\mathbb{R}^{D\times P}$, which costs
$\mathcal{O}(DP)$ per sample and $\mathcal{O}(NDP)$ over the
training set. In a batched implementation, it must also hold full parameter
gradients and the projection matrix, incurring $\mathcal{O}(BD)$ and
$\mathcal{O}(DP)$ memory, respectively, for batch size $B$. This makes
projected gradient construction memory-bound for billion-parameter VLA models.

Following LoGRA~\cite{choe2026logra}, we avoid explicit gradient projection in parameter space by exploiting the Kronecker structure of linear layer gradients. Specifically, for layer $\ell$ with activation $x_i^\ell$ and backpropagated error $\delta_i^\ell$, the weight gradient admits the outer product form $\delta_i^\ell(x_i^\ell)^\top$. ATHENA projects these two factors before forming the projected feature:
\begin{equation}
    \bar{g}_i^\ell =
    (P_{\mathrm{out}}^\ell)^\top
    \delta_i^\ell
    (x_i^\ell)^\top
    P_{\mathrm{in}}^\ell .
    \label{eq:weigh_gradient}
\end{equation}
The resulting layer features are flattened and concatenated across layers. This replaces dense projection in parameter space with bilateral projection in activation space, reducing the leading projection cost from $\mathcal{O}(DP)$ to $\mathcal{O}(\sqrt{DP})$ while avoiding materialization of full gradients in
$\mathbb{R}^{D}$.

\paragraph{Random Truncated Approximation.}
After Kronecker compressed featurization, ATHENA stacks the timestep features into $G\in\mathbb{R}^{N\times P}$, where each row is $\phi_i^\top$. Following Eq.~\eqref{eq:trak_score}, a dense damped Gauss--Newton approximation requires forming and inverting $G^\top G+\lambda I_P$, which costs
$\mathcal{O}(NP^2+P^3)$. This cost grows quickly with the number of timesteps, demonstrations, and tasks.

We avoid this dense projected inverse by applying a rank-$r$ Random Truncated Approximation (RTA)~\cite{2011randomSVD, nakatsukasa2020fast_randomsvd} to $G$. Concretely, ATHENA approximates the compressed gradient matrix by its leading randomized spectral components,
$G\approx U_r\Sigma_rV_r^\top$, where $V_r\in\mathbb{R}^{P\times r}$ and $r\ll P$. Since $G^\top G\approx V_r\Sigma_r^2V_r^\top$, ATHENA projects each timestep feature into the retained subspace as $\tilde{\phi}_i=V_r^\top\phi_i$, and computes:
\begin{equation}
    \widehat{\psi}_{\mathrm{RTA}}(z_{\mathrm{te}},z_{\mathrm{tr}})
    =
    \tilde{\phi}_{\mathrm{te}}^\top
    (\Sigma_r^2+\lambda I_r)^{-1}
    \tilde{\phi}_{\mathrm{tr}} .
    \label{eq:rank_r}
\end{equation}
This reduces the leading data-dependent cost from
$\mathcal{O}(NP^2)$ to
$\mathcal{O}(NPr)$, while replacing a $P$-dimensional dense
inverse with an $r$-dimensional damped inverse. Additional details on Kronecker compressed featurization and RTA are provided in Appendix~\ref{app:complexity}.

\subsection{Interactive Influence for Multitask Scaling}
\label{sec:mii}

In multitask VLA curation, each demonstration may affect its own task differently from other tasks. To capture these interactions, we propose Multitask Influence Interaction (MII), which models both the influence on the demonstration's own task and other tasks to enable balanced multitask data curation.
Let demonstration $i$ belong to task $c(i)\in\{1,\dots,K\}$, with evaluation rollouts $\mathcal{E}$. We define task-local and cross-task influence as:
\begin{equation}
    \widetilde{\Psi}^{c(i)}_{\pi\text{-inf}}(i) = \frac{1}{H_i}\sum_{\tau_j \in \mathcal{E}_{c(i)}} R(\tau_j) S_{j,i}, \quad
    \widetilde{\Psi}^{\mathrm{all}-c(i)}_{\pi\text{-inf}}(i) = \frac{1}{H_i}\sum_{\tau_j \in \mathcal{E}\setminus \mathcal{E}_{c(i)}} R(\tau_j) S_{j,i},
\end{equation}
where $S_{j,i}$ denotes the contribution of demonstration $i$ to rollout $\tau_j$ (aggregated over timestep pairs). The first term measures local influence, while the second captures cross-task effects.

Raw values of $\widetilde{\Psi}^{c(i)}_{\pi\text{-inf}}$ and $\widetilde{\Psi}^{\mathrm{all}-c(i)}_{\pi\text{-inf}}$ are defined on separate rollout sets and have different numerical scales. We convert each component to a normalized, sorted score to combine local and cross-task influence consistently in the MII score.

\begin{equation}
    r_i^{c(i)} = \operatorname{rank}_{\downarrow}^{c(i)}\bigl(\widetilde{\Psi}^{c(i)}_{\pi\text{-inf}}(i)\bigr), \quad
    r_i^{\mathrm{all}-c(i)} = \operatorname{rank}_{\downarrow}^{\mathrm{all}-c(i)}\bigl(\widetilde{\Psi}^{\mathrm{all}-c(i)}_{\pi\text{-inf}}(i)\bigr),
\end{equation}
and define normalized utilities:
\begin{equation}
    u_i^{c(i)} = \max(\varepsilon, 1 - r_i^{c(i)}/n_{c(i)}), \quad
    u_i^{\mathrm{all}-c(i)} = \max(\varepsilon, 1 - r_i^{\mathrm{all}-c(i)}/(n-n_{c(i)})),
\end{equation}
where $n_{c(i)}$ is the number of demonstrations in task $c(i)$, $n$ is the total number of demonstrations, and $\varepsilon>0$ is a small numerical floor.

\begin{proposition}[MII]
\label{pro:MII}
The cross-task influence function for curating balanced subsets of demonstrations from multitask datasets is mathematically formulated as:
\begin{equation}
    f_i^{\mathrm{MII}} = u_i^{c(i)} \cdot u_i^{\mathrm{all}-c(i)}.
\end{equation}

Training demonstrations are sorted by $f_i^{\mathrm{MII}}$ to curate a balanced subset that preserves locally critical examples while accounting for cross-task interactions.
\end{proposition} 

The details of the multitask estimation procedure are summarized in Appendix~\ref{app:A.3}.

\section{Experiments}
\label{sec:result}

\subsection{Experimental Setup}
\label{sec:exp_setup}

We evaluate ATHENA on RoboTwin 2.0~\cite{mu2024robotwin} and six real robot manipulation tasks, using 50K training steps as in CUPID~\cite{agia2025cupid}, comparing against Oracle~\cite{mandlekar2021robomimic}, Random~\cite{agia2025cupid}, TAROT~\cite{pmlr-v267-feng25l}, TSS~\cite{yang2026less}, and Distillation~\cite{chen2025ftncfm}. Baselines, implementation,  and evaluation details are provided in Appendix~\ref{app:B}. 

\subsection{Data Curation for Billion-Parameter Multitask VLAs}
\label{sec:Simulation}

\subsubsection{RoboTwin 2.0 Simulation Benchmark}
\label{sec:robotwin_50_benchmark}
\begin{wrapfigure}{r}{0.43\linewidth}
    \vspace{-0.7em}
    \centering
    \includegraphics[width=\linewidth]{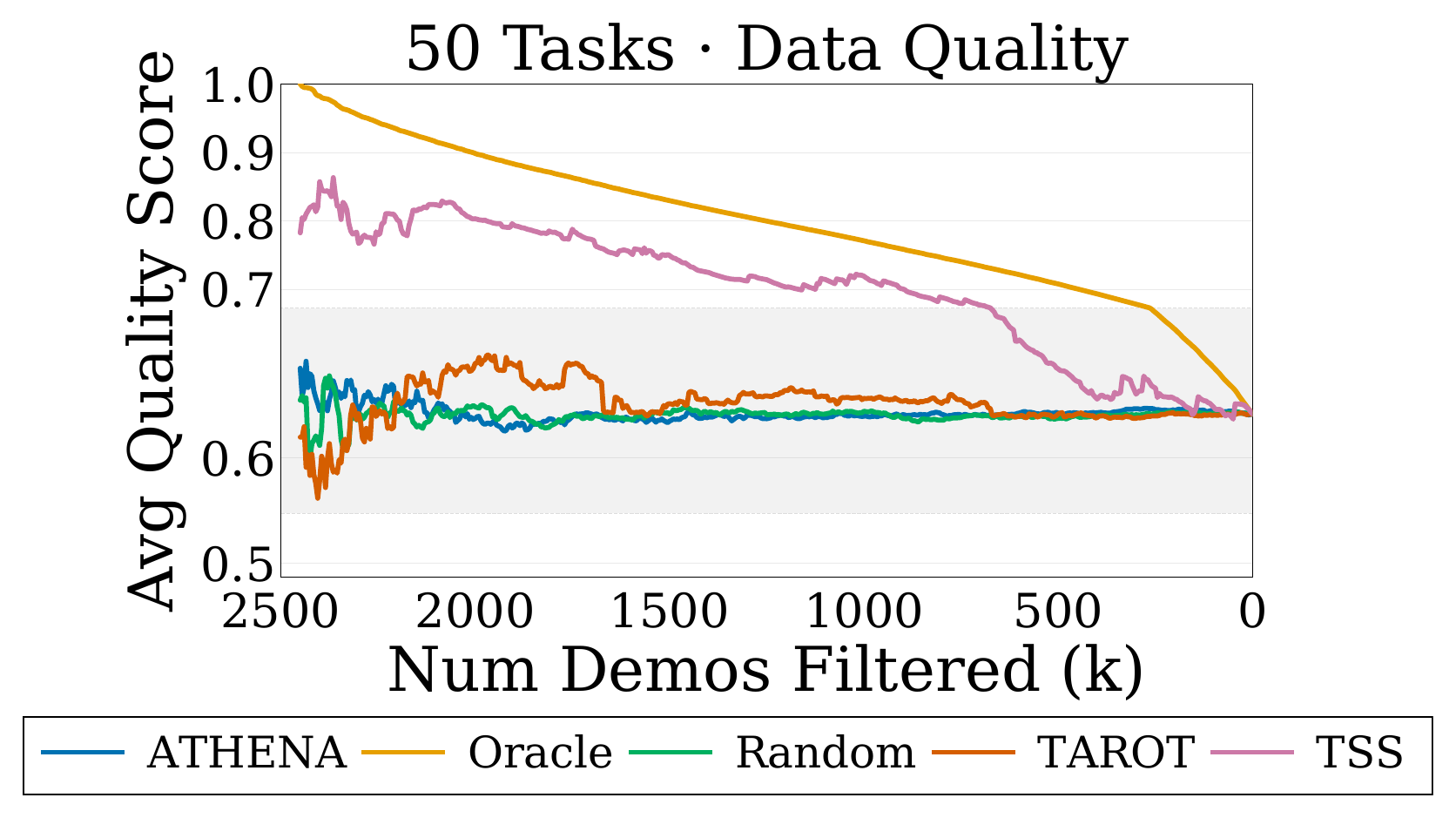}
    \vspace{-1.5em}
    \caption{{Quality scores over 50 tasks.} Shaded band shows a $3\times$ vertical zoom to reveal score variations.}
    \label{fig:quality_only_50task}
    \vspace{-0.6em}
\end{wrapfigure}
Following the official RoboTwin 2.0 protocol, we curate and fine-tune on the demo\_clean split, which contains 2,500 demonstrations across 50 tasks totaling 9.34 hours at 16.67 Hz, and evaluate under clean and randomized generalization settings.

\begin{figure}[t]
    \centering
    \includegraphics[width=\linewidth]{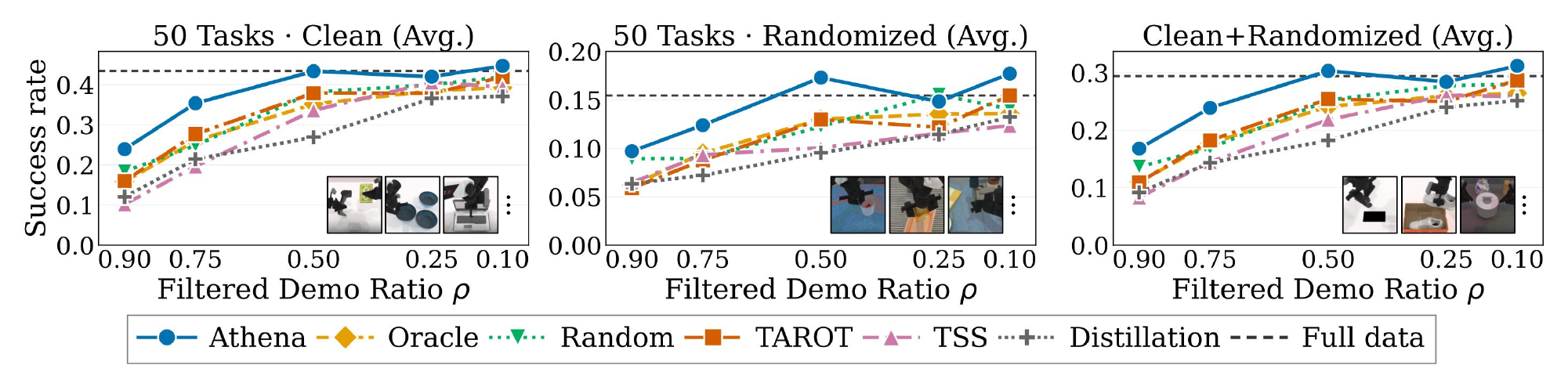}
    \vspace{-0.6cm}
    \caption{ \textbf{Average success rate over $K=50$ tasks.} Results across $\rho$ under clean and randomized evaluation settings; the right panel reports their mean. Dashed lines denote full-data fine-tuning.
} 
    \label{fig:multitask_quality_success_50}
\end{figure}

Figures~\ref{fig:quality_only_50task} and~\ref{fig:multitask_quality_success_50} compare various methods in terms of 50-task quality scores~\cite{agia2025cupid,mandlekar2021robomimic} and fine-tuning success rates across different data budgets, respectively. The quality score of the distillation-based method is omitted because it does not perform demonstration-level curation, while its success rate remains low. Although Oracle and TSS retain subsets with higher quality scores, both yield suboptimal success rates.  
Conversely, despite exhibiting less prominent quality scores, ATHENA reaches 44.70\% clean and 17.72\% randomized success at $\rho=0.1$, outperforming full-data training (43.42\% and 15.44\%). Notably, with only 50\% of demonstrations retained, ATHENA still matches full-data training under the clean evaluator (43.36\% vs. 43.42\%) and exceeds it under the randomized evaluator (17.30\% vs. 15.44\%), yielding a 0.90-point higher average success rate (30.33\% vs. 29.43\%), corresponding to a cumulative 45.0-point improvement across the 50 tasks. More importantly, even this 50\%-data setting remains close to the single-task $\pi_0$ fine-tuning performance reported by RoboTwin 2.0~\cite{mu2024robotwin} under clean evaluation (43.36\% vs. 46.42\%) and exceeds it under randomized evaluation (17.30\% vs. 16.34\%). Unlike single-task training, which requires one fine-tuning and checkpoint per task, ATHENA uses a joint fine-tuning paradigm, reducing computation and storage by tens of times. These contrasting results demonstrate that expert-defined quality is not necessarily predictive of downstream success and highlight the high ROI of ATHENA in multitask VLA fine-tuning.

Furthermore, TAROT achieves moderate gains using Whitened Feature Distance for multitask balancing, but its dependence on precise target-set construction limits further improvements. The Random method also brings improvements, but its uniform retention strategy relies heavily on the quality distribution of the original dataset. Per-task success rates are provided in Appendix~\ref{app:appendix_per_task}. Overall, ATHENA demonstrates superior efficacy in curating data for VLA fine-tuning in simulation.

\begin{figure}[t]
    \centering
    \includegraphics[width=\linewidth]{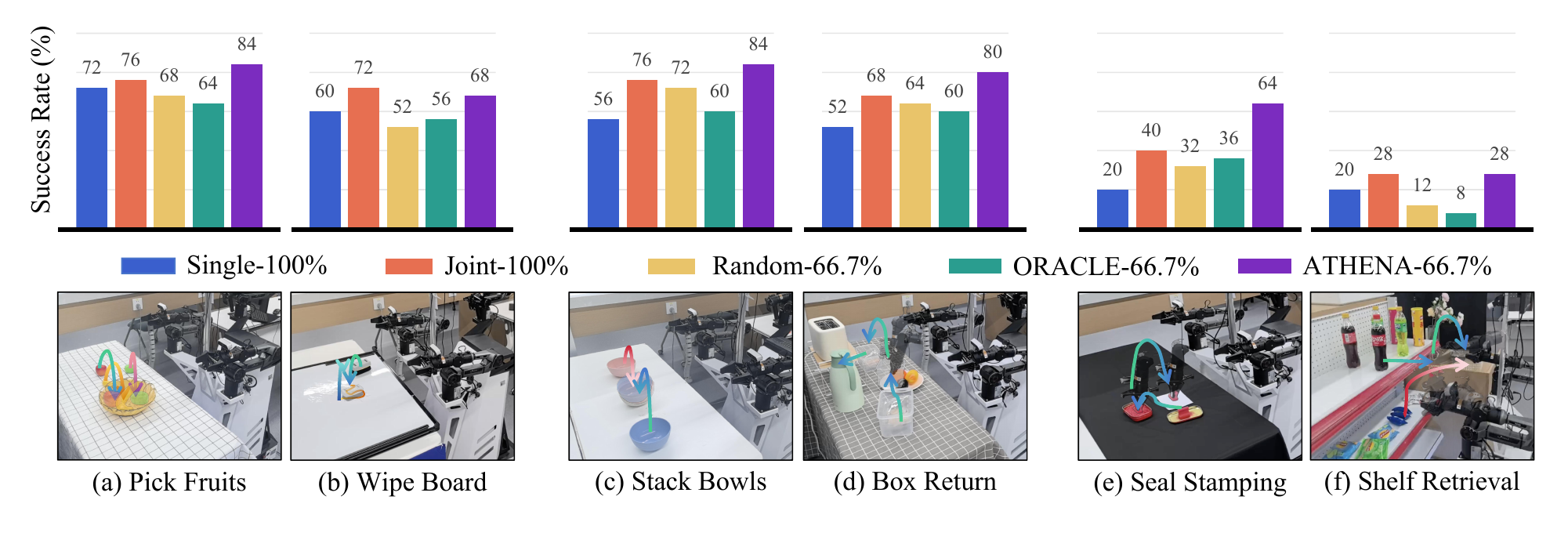}
    \vspace{-1.0cm}
    \caption{
        \textbf{Real robot evaluation ($K=6$).}
        Success rates over 25 trials across six ALOHA tasks of varying difficulty, compared with representative baselines.
    }
    \label{fig:real_robot_success1}
\end{figure}

\subsubsection{Real Robot Experiments}
To further evaluate ATHENA, we conduct experiments across six real-robot ALOHA tasks spanning three difficulty levels, comprising 720 high-quality demonstrations totaling 6.9 hours collected at 25 Hz. The suite comprises two simple tasks (Pick Fruits, Wipe Board), two medium tasks (Stack Bowls, Box Return), and two challenging long-horizon tasks (Seal Stamping, Shelf Retrieval). To assess positional generalization, we perform 25 trials per task with randomized object positions. Detailed task setups are deferred to Appendix~\ref{app:B.2}. 

As shown in Figure~\ref{fig:real_robot_success1}, the Single-100\% baseline requires six independent fine-tuning checkpoints (totaling 300K training steps and 240GB of storage), yet yields only 46.7\% average success. The Joint-100\% baseline consolidates this into a single fine-tuning run, drastically reducing these resource requirements while improving average success to 60.0\%, underscoring the necessity and high ROI of joint training. However, it still struggles with negative-influence data. Random-66.7\% and Oracle-66.7\% attempt to filter out these data, but instead drop success rates to 50.0\% and 47.3\%, respectively. In contrast, ATHENA successfully filters out low-utility data while preserving essential multitask knowledge, achieving the highest average success of 68.0\% with only 66.7\% of the data, corresponding to cumulative gains of 48.0 points over Joint-100\% and 128 points over Single-100\% across the six tasks. These results demonstrate ATHENA's superior real-robot performance and ROI.

\section{Ablation and Discussion}
\label{sec:discussion}
\subsection{Ablation of Computational Efficiency}
\label{sec:logra_ablation}

To validate the contributions of Kronecker-compressed featurization and Random Truncated Approximation (RTA) to scalable influence computation, we perform a computational ablation by removing accelerated components from ATHENA. This ablated variant reduces to a standard projected-gradient influence pipeline, where per-sample gradients are explicitly projected, and the projected Gauss--Newton matrix is inverted densely. 
This unoptimized pipeline matches the computational form of TRAK-style~\cite{park2023trak} and CUPID-style~\cite{agia2025cupid} influence estimation, and therefore serves as a natural reference for evaluating ATHENA's computational speedup.
Table~\ref{tab:scalability_time} reports the total computation time across varying task scales (\(K \in \{5, 10, 25, 50\}\), 30.2K--560.5K demonstration timesteps) on 140\,GB GPUs. ATHENA achieves a speedup of up to \(313.4\times\) over this unoptimized baseline. A detailed per-component breakdown is provided in Appendix~\ref{app:C.1}.

\begin{table}[h]
    \centering
    \footnotesize 
    \setlength{\tabcolsep}{8pt} 
    
    \caption{\textbf{Computation time (GPU-hours) under data scaling.} Columns denote task counts ($K$) and corresponding demonstration timesteps.}
    \label{tab:scalability_time}
    
    \begin{tabular}{lcccc}
        \toprule
        \multirow{2}{*}{Method} 
        & $K=5$ 
        & $K=10$ 
        & $K=25$ 
        & $K=50$ \\
        & \scriptsize{(30.2K steps)} 
        & \scriptsize{(66.5K steps)} 
        & \scriptsize{(291.9K steps)} 
        & \scriptsize{(560.5K steps)} \\
        \midrule
        
        w/o Acceleration 
        & $\sim 446.2$ 
        & $\sim 885.5$ 
        & $\sim 3297.4$  
        & $\sim 8054.6$  \\
        
        ATHENA
        & $\mathbf{\sim 1.1}$ 
        & $\mathbf{\sim 2.4}$ 
        & $\mathbf{\sim 14.0}$ 
        & $\mathbf{\sim 25.7}$  \\
        \midrule
        
        Speedup 
        & $\mathbf{\sim 405.6\times}$ 
        & $\mathbf{\sim 369.0\times}$ 
        & $\mathbf{\sim 235.5\times}$ 
        & $\mathbf{\sim 313.4\times}$  \\
        \bottomrule
    \end{tabular}
\end{table}

\subsection{Multitask Joint Fine-tuning Ablation}
\label{sec:6.2}
\begin{figure}[h]
    \centering
    \includegraphics[width=\linewidth]{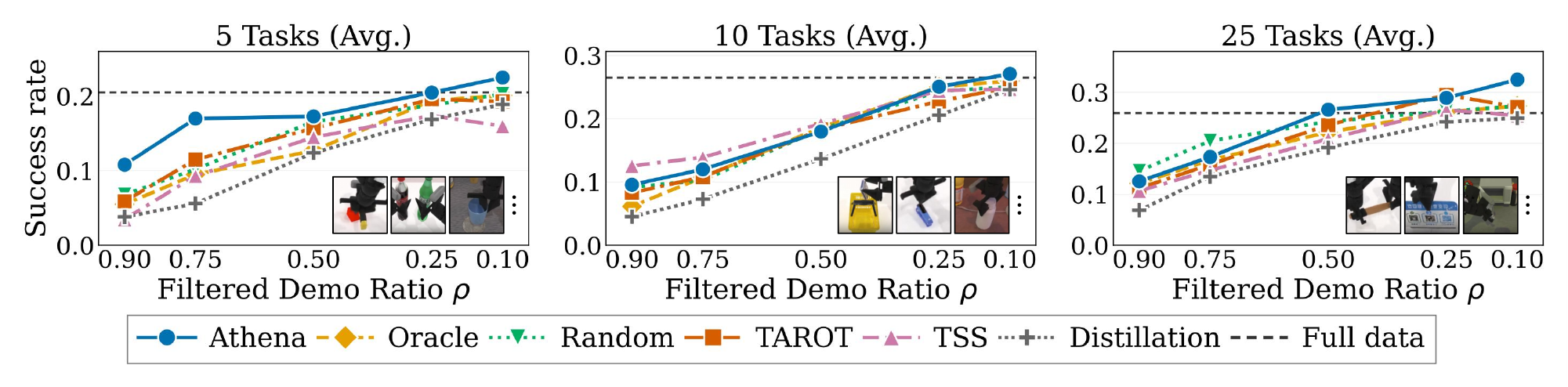}
    \vspace{-0.6cm}
\caption{Mean success rates for varying task counts under clean and randomized evaluations.}
    \label{fig:5}
\end{figure}

To ablate ATHENA under varying task counts, we evaluate $K\in\{5,10,25\}$ task subsets sampled from the full RoboTwin $50$-task benchmark. As shown in Figure~\ref{fig:5}, at $\rho=0.1$, ATHENA outperforms other baselines across all three task scales, with the largest improvement at $K=25$. In this setting, ATHENA achieves 32.48\% mean success over both evaluation settings, compared with 25.93\% for full-data fine-tuning. Detailed per-setting results are reported in Appendix~\ref{app:multitask_results}. At $\rho=0.5$, the absolute data scale becomes a limiting factor: all methods at $K=5$ and $K=10$ remain below full-data fine-tuning, whereas $K=25$ provides a larger data pool for curation to be effective. This suggests that data curation yields higher returns as data scale increases. Appendix~\ref{app:MII} further analyzes retention imbalance without MII and reports additional single-task curation studies, including real-robot cases on multi-peak actions and spurious correlations.

\subsection{Curation Under Mixed Clean-Randomized Data}

We next ask whether ATHENA remains effective beyond the clean setting used above. Following the RoboTwin 2.0 clean and randomized (Rand.) configurations, we build a three-task mixed pool and compare it with a compact clean control. 

\begin{wrapfigure}[10]{r}{0.42\linewidth}
    \vspace{-1.4em}
    \centering
    \includegraphics[width=\linewidth]{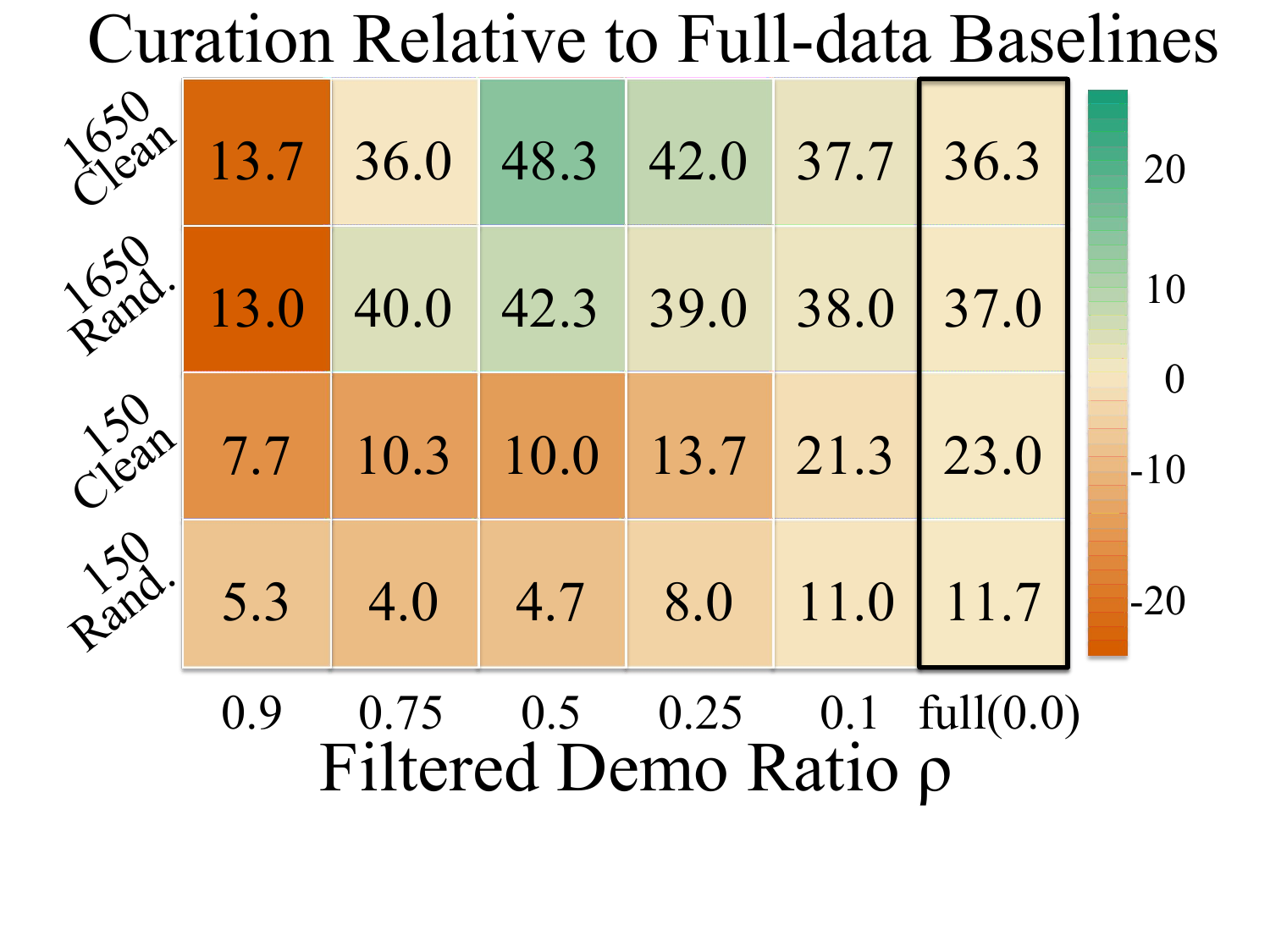}
    \vspace{-3em}
    \caption{
    \textbf{Curation with mixed data.}
    }
    \label{fig:three_task_clean_random}
    \vspace{-1.0em}
\end{wrapfigure}

As shown in Figure~\ref{fig:three_task_clean_random}, the 150 demo clean pool yields only
limited variation across filtering ratios, since the small-scale dataset has low
redundancy and leaves little room for subset optimization. In contrast, scaling to the 1650 demo clean and random pool reveals a clear benefit from curation. At $\rho=0.50$, ATHENA improves average success from 36.3\% to 48.3\% under clean evaluation and from 37.0\% to 42.3\% under randomized evaluation. These results show that ATHENA extracts high ROI subsets from heterogeneous fine-tuning data.

\subsection{Cross VLA Model Transferability of Curated Data}
\begin{wraptable}[11]{r}{0.54\linewidth}
    \vspace{-0.8em}
    \centering
    \footnotesize 
    \setlength{\tabcolsep}{4.0pt} 
    \renewcommand{\arraystretch}{1.1} 
    
    \caption{\textbf{Cross VLA transfer.} Success rates ($\%$, $\uparrow$) for $\pi_{0.5}$ using subsets curated on $\pi_0$.}
    \label{tab:pi0_guides_pi05}
    \vspace{0.2em}

    \begin{tabular}{l cccccc} 
        \toprule
        \multirow{2}{*}{Eval} & \multicolumn{5}{c}{Filtered Demo Ratio $\rho$} & \multirow{2}{*}{Full} \\
        \cmidrule(lr){2-6}
                              & 0.90 & 0.75 & 0.50 & 0.25 & 0.10 & 0.0 \\
        \midrule
        Clean & 37.86 & 53.92 & 63.28 & 64.16 & 67.30 & 57.00 \\
        Rand. & 16.80 & 26.74 & 34.23 & 37.16 & 37.77 & 25.68 \\
        AVG   & 27.33 & 40.33 & 48.76 & 50.66 & 52.54 & 41.34 \\
        \bottomrule
    \end{tabular}
\end{wraptable}
Table~\ref{tab:pi0_guides_pi05} illustrates that robot demonstration subsets curated via ATHENA on $\pi_0$ substantially enhance $\pi_{0.5}$ performance~\cite{black2024pi0,intelligence2025pi_05}, increasing the average success rate across 50 tasks. Both clean and randomized conditions yield average success rates significantly above full-data baselines, indicating that high-influence demonstrations identified by ATHENA generalize across VLA model variants. Detailed per-task success metrics are provided in Appendix~\ref{app:appendix_per_task}.

\section{Conclusion}
\label{sec:conclusion}
In this work, we present ATHENA, an influence function framework tailored for multitask VLA data curation at the billion-parameter scale. To scale influence estimation to VLA models, ATHENA incorporates Kronecker-structured linear-layer gradients and a rank-$r$ Random Truncated Approximation. Furthermore, it formulates Multitask Influence Interaction to balance multitask data curation and improve ROI. Experiments on the 50-task RoboTwin 2.0 benchmark and six real-robot tasks show that ATHENA outperforms full-data fine-tuning and competitive baselines while using fewer demonstrations, indicating its effectiveness for data curation in VLA fine-tuning.

\section{Limitations}

\textbf{Pretraining data curation.}
ATHENA focuses on VLA fine-tuning, while model performance also depends on pretraining with more heterogeneous data across tasks, embodiments, and distributions~\cite{black2024pi0,lingbo-vla,lingbo-va,univla}. Influence estimates from fine-tuning may not directly transfer to this setting, and extending ATHENA to robot VLA pretraining remains future work.

\textbf{Real-robot rollout cost.}
Influence-based robot curation currently requires evaluation rollouts to estimate downstream effects~\cite{koh2017understanding,agia2025cupid,lee2026quality_cupid2}. However, collecting rollouts on physical robots for each task incurs substantially higher cost than in simulation. Scaling to broader real-world settings requires more efficient rollout protocols or lower-cost proxies for downstream robot performance.

\bibliography{example}


\appendix

\clearpage
\section*{Appendix}
\section{Additional Implementation Details}
\label{app:A}

\subsection{Action-level Influence for Flow-matching VLAs}
\label{app:action_influence}

Section~\ref{sec:if_curation} introduces $\Psi_{a\text{-inf}}$ in Eq.~\eqref{eq:perf_inf} as the action-level
term for closed-loop performance influence. Referring to the closed-loop
attribution principle of CUPID~\cite{agia2025cupid}, demonstration-level
performance influence is decomposed into action-level influences evaluated on
state-action pairs from closed-loop rollouts. In our setting, Eq.~\eqref{eq:perf_inf} retains
this decomposition, while the action-level influence is derived for the
flow-matching action parameterization of $\pi_0$.

\paragraph{Flow-matching action objective.}
Given a state-action pair $z=(s,a)$, a flow-matching VLA perturbs the action
along the linear path
\begin{equation}
x_t = t\epsilon + (1-t)a, \qquad \epsilon \sim \mathcal{N}(0,I), \quad t\in[0,1],
\end{equation}
where $x_0=a$ and $x_1=\epsilon$. The policy is trained to predict the velocity
field along this path. Since
\begin{equation}
\frac{d x_t}{dt} = \epsilon-a,
\end{equation}
the supervised flow-matching loss is
\begin{equation}
loss_{\mathrm{flow}}(s,a;\theta)
=
\mathbb{E}_{t,\epsilon}
\left[
\left\|v_\theta(x_t,s,t)-(\epsilon-a)\right\|^2
\right].
\label{eq:app_flow_loss}
\end{equation}
This objective defines the fine-tuning loss of the flow-matching policy.
\paragraph{From action likelihood to a flow surrogate.}
The original action influence in CUPID measures how a training state-action pair
$(s,a)$ affects the likelihood of a rollout action $(s',a')$ under the learned
policy. In likelihood-based policies, this can be written as an influence on
$\log \pi_\theta(a'|s')$. However, for diffusion or flow-matching policies,
directly evaluating and differentiating $\log \pi_\theta(a'|s')$ is nontrivial
because actions are generated through an iterative generative process rather
than a closed-form density. We therefore define a scalar flow-matching surrogate
as the rollout-side test quantity $f(\hat z;\theta)$ in Eq.~\eqref{eq:if_classic}.

For flow-matching VLAs, we define the square-flow attribution surrogate as
\begin{equation}
f_{\mathrm{sf}}(s,a;\theta)
=
\mathbb{E}_{t,\epsilon}
\left[
\|v_\theta(x_t,s,t)\|^2
\right].
\label{eq:app_square_flow}
\end{equation}
Here $x_t=t\epsilon+(1-t)a$, with $\epsilon\sim\mathcal{N}(0,I)$ and $t\in[0,1]$.
This quantity measures the magnitude of the model's velocity response around
the queried state-action pair. Unlike the supervised flow-matching loss in
Eq.~\eqref{eq:app_flow_loss}, it does not require an expert velocity label on
rollout samples and maps the high-dimensional action velocity prediction to a
scalar attribution target.

\paragraph{Action-level influence.}
Let $z=(s,a)$ be a training state-action pair and
$\hat z=(\hat s,\hat a)$ be a rollout state-action pair. Following the
first-order influence form in Eq.~\eqref{eq:if_classic}, we derive the flow-matching action
influence as
\begin{equation}
\Psi_{a\text{-inf}}(\hat z,z)
=
-
\nabla_\theta f_{\mathrm{sf}}(\hat z;\theta^\star)^\top
H_\theta^{-1}
\nabla_\theta \mathcal{L}(z;\theta^\star),
\label{eq:app_action_inf_sf}
\end{equation}
where $\theta^\star$ denotes the fine-tuned VLA model parameters, $L$ is the
imitation loss in Section~\ref{sec:if_curation}, and $H_\theta=\nabla_\theta^2 \mathcal{ L}(z_{t}^{i};
\theta^\star)$ follows the notation of Eq.~\eqref{eq:if_classic}. For the flow-matching VLA model
$\pi_0$, $\mathcal{L}(z;\theta^\star)$ is given by the supervised flow-matching objective
in Eq.~\eqref{eq:app_flow_loss}. In practice, we do not form $H_\theta^{-1}$
explicitly. ATHENA computes this interaction in the projected gradient space
using the Kronecker-compressed features and the rank-$r$ Random Truncated
Approximation described in Section~\ref{sec:athena_method} and Appendix~\ref{app:complexity}.

\subsection{Complexity Analysis}
\label{app:complexity}

\paragraph{Unaccelerated influence estimation complexity.}
We first derive the time complexity of the main computational bottlenecks in
Sections~\ref{sec:prelim} and~\ref{sec:athena_method}. For multiplying
$A\in\mathbb{R}^{a\times b}$ and $B\in\mathbb{R}^{b\times c}$ using standard
matrix multiplication, the time complexity is $\mathcal{O}(abc)$. Thus, the
projected-gradient operation in Eq.~\eqref{eq:projected_grad}, $\Omega^\top g_i$ with
$\Omega^\top\in\mathbb{R}^{P\times D}$ and $g_i\in\mathbb{R}^{D}$, requires
$\mathcal{O}(P\cdot D\cdot 1)=\mathcal{O}(DP)$ time per timestep, and
$\mathcal{O}(NDP)$ time over $N$ timesteps.

Similarly, forming $G^\top G$ in Eq.~\eqref{eq:trak_score} multiplies
$G^\top\in\mathbb{R}^{P\times N}$ by $G\in\mathbb{R}^{N\times P}$, requiring
$\mathcal{O}(P\cdot N\cdot P)=\mathcal{O}(NP^2)$ operations. The result is a
$P\times P$ matrix, whose inversion requires $\mathcal{O}(P^3)$ operations.
Therefore, the dense projected inverse-Hessian stage has computational
complexity $\mathcal{O}(NP^2+P^3)$. For the experimental setting used in this
paper, the $\pi_0$ model has $D=3.3$B parameters, with projected dimension
$P=4096$ and up to $N=560.5$K demonstration timesteps. At this scale, the
unaccelerated projected influence pipeline requires approximately $8054.6$ GPU-hours at
$K=50$, as reported in Table~\ref{tab:scalability_time}, highlighting the
computational burden of the original $\mathcal{O}(NDP)$ and
$\mathcal{O}(NP^2+P^3)$ pipeline.

\paragraph{Kronecker compressed gradient featurization.}
We expand the Kronecker-structured projection used in Eq.~\eqref{eq:weigh_gradient}. For a linear
layer $\ell$ such as an MLP layer, let
$W^\ell\in\mathbb{R}^{d_\ell^{\rm out}\times d_\ell^{\rm in}}$, with layer
parameter dimension $D_\ell=d_\ell^{\rm out}d_\ell^{\rm in}$. For timestep
$i$, the input activation and backpropagated error are
$x_i^\ell\in\mathbb{R}^{d_\ell^{\rm in}}$ and
$\delta_i^\ell\in\mathbb{R}^{d_\ell^{\rm out}}$, respectively. The per-timestep weight gradient is
\begin{equation}
    G_i^\ell
    =
    \nabla_{W^\ell}L(z_i;\theta^\star)
    =
    \delta_i^\ell (x_i^\ell)^\top .
\end{equation}
After vectorization, using $\mathrm{vec}(ab^\top)=b\otimes a$, where
$\otimes$ denotes the Kronecker product, we obtain
$g_i^\ell=\mathrm{vec}(G_i^\ell)=x_i^\ell\otimes\delta_i^\ell$.

A dense projection applies $\Omega_\ell^\top g_i^\ell$ with
$\Omega_\ell\in\mathbb{R}^{D_\ell\times P}$, where $P$ is the projected
gradient dimension in Eq.~\eqref{eq:projected_grad}, requiring $\mathcal{O}(D_\ell P)$ operations
for layer $\ell$. Following LoGRA~\cite{choe2026logra}, ATHENA imposes a
Kronecker-product structure on the projection matrix:
\begin{equation}
    \Omega_\ell
    =
    P_{\rm in}^\ell\otimes P_{\rm out}^\ell .
\end{equation}
By the mixed-product property of Kronecker products,
\begin{equation}
    \Omega_\ell^\top g_i^\ell
    =
    \bigl((P_{\rm in}^\ell)^\top x_i^\ell\bigr)
    \otimes
    \bigl((P_{\rm out}^\ell)^\top\delta_i^\ell\bigr).
\end{equation}
Equivalently, the same projected feature can be computed in matrix form as
\begin{equation}
    \bar{g}_i^\ell
    =
    (P_{\rm out}^\ell)^\top
    \delta_i^\ell (x_i^\ell)^\top
    P_{\rm in}^\ell ,
\end{equation}
which is the compact bilateral projection in Eq.~\eqref{eq:weigh_gradient}.

This bilateral form avoids materializing the full layer gradient
$g_i^\ell$ and performs projection directly in activation space. When
$d_\ell^{\rm in}\approx d_\ell^{\rm out}\approx\sqrt{D_\ell}$ and the two
projected activation dimensions are both on the order of $\sqrt{P}$, the
per-layer projection cost reduces from $\mathcal{O}(D_\ell P)$ to
$\mathcal{O}(\sqrt{D_\ell P})$~\cite{choe2026logra}. Applying this reduction across projected
layers and $N$ timesteps give the overall projected-gradient construction
cost of $\mathcal{O}(N\sqrt{DP})$, compared with the dense
$\mathcal{O}(NDP)$ cost.

\paragraph{Random Truncated Approximation.}
After Kronecker compressed featurization, ATHENA stacks the projected timestep
features into $G\in\mathbb{R}^{N\times P}$, where each row is $\phi_i^\top$.
Following Eq.~\eqref{eq:trak_score}, the dense projected inverse requires forming and inverting
$G^\top G+\lambda I_P$, with complexity
$\mathcal{O}(NP^2+P^3)$.

To avoid this dense projected inverse, ATHENA applies a rank-$r$ Random
Truncated Approximation (RTA)~\cite{2011randomSVD,nakatsukasa2020fast_randomsvd}
to $G$. Specifically, we approximate
$G\approx U_r\Sigma_r V_r^\top$, where $V_r\in\mathbb{R}^{P\times r}$ and
$r\ll P$, so that $G^\top G\approx V_r\Sigma_r^2V_r^\top$. Each projected feature is then mapped to the retained subspace as
$\tilde{\phi}_i=V_r^\top\phi_i$, leading to the rank-$r$ influence surrogate
in Eq.~\eqref{eq:rank_r}:
\[
    \widehat{\psi}_{\rm RTA}(z_{\rm te},z_{\rm tr})
    =
    \tilde{\phi}_{\rm te}^{\top}
    (\Sigma_r^2+\lambda I_r)^{-1}
    \tilde{\phi}_{\rm tr}.
\]

The randomized range-finding step and the projection of all $N$ timestep
features into the retained subspace both scale as $\mathcal{O}(NPr)$ up to
lower-order terms. Therefore, RTA reduces the projected inverse-Hessian stage
from the dense $\mathcal{O}(NP^2+P^3)$ computation to
$\mathcal{O}(NPr)$.

\subsection{Multitask data curation }
\label{app:A.3}

Algorithm~\ref{alg:mii} summarizes the computation of the Multitask Influence Interaction (MII) score introduced in Section~\ref{sec:athena_method}. For each demonstration, MII separately estimates its task-local utility from its own task and its cross-task utility over rollouts from all other tasks. The two utilities are then rank-normalized within each task group and combined to obtain the final curation score. When $K=1$, no other tasks exist, and the procedure reduces to
standard single-task influence-based data curation. Single-task simulation and real-robot results are provided in Appendix~\ref{app:MII} and Figs.~\ref{fig:single_task_if}, \ref{fig:stack_bowls_multipeak}, and \ref{fig:box_return_spurious}.

\begin{algorithm}[t]
\caption{Multitask Influence Interaction (MII) Score Computation}
\label{alg:mii}
\begin{algorithmic}[1]

\State \textbf{Input:} Demonstration set $\mathcal{D}=\{i\}_{i=1}^n$, task index $c(i)$, rollout set $\mathcal{E}$, reward function $R(\cdot)$, influence score $S_{j,i}$

\State Partition rollouts into task-specific sets $\{\mathcal{E}_k\}_{k=1}^K$

\For{each demonstration $i \in \mathcal{D}$}

    \State Compute task-local influence:
    \[
    \widetilde{\Psi}^{c(i)}_{\pi\text{-inf}}(i)
    = \frac{1}{H_i}\sum_{\tau_j \in \mathcal{E}_{c(i)}} R(\tau_j) S_{j,i}
    \]

    \State Compute cross-task influence:
    \[
    \widetilde{\Psi}^{\mathrm{all}-c(i)}_{\pi\text{-inf}}(i)
    = \frac{1}{H_i}\sum_{\tau_j \in \mathcal{E}\setminus \mathcal{E}_{c(i)}} R(\tau_j) S_{j,i}
    \]

\EndFor

\State Compute ranks within each task group:
\[
r_i^{k} = \operatorname{rank}_{\downarrow}^{k}(\cdot)
\]

\State Normalize utilities:
\[
u_i^{c(i)},\; u_i^{\mathrm{all}-c(i)}
\]

\State Compute final score:
\[
f_i^{\mathrm{MII}} = u_i^{c(i)} \cdot u_i^{\mathrm{all}-c(i)}
\]

\State \textbf{Output:} ranked demonstrations by $f_i^{\mathrm{MII}}$

\end{algorithmic}
\end{algorithm}

\section{Experimental Details}
\label{app:B}

This section collects extended experimental specifications referenced from Section~\ref{sec:result}.

\subsection{RoboTwin 2.0 Simulation Benchmark}
\label{app:B.1}

We conduct simulation experiments on RoboTwin 2.0 using its official simulator, task definitions, success criteria, and rollout evaluation protocol. We use the $K{=}50$ task suite to evaluate large-scale multitask VLA fine-tuning and data curation. We report results under two RoboTwin 2.0 evaluation configurations. The demo\_clean configuration corresponds to the Easy setting, while demo\_randomized corresponds to the Hard setting with stronger scene randomization and more challenging initial conditions~\cite{mu2024robotwin}.

\subsection{Real-Robot Task Setups}
\label{app:B.2}

\begin{table*}[!t]
\centering
\caption{Descriptions of the six real-robot manipulation tasks, each with 120 demonstrations.}
\label{tab:appendix_task_descriptions}

\renewcommand{\arraystretch}{1.15}
\setlength{\tabcolsep}{4pt}
\footnotesize

\begin{tabular}{
>{\centering\arraybackslash}m{0.20\textwidth}
>{\raggedright\arraybackslash}m{0.28\textwidth}
>{\centering\arraybackslash}m{0.20\textwidth}
>{\raggedright\arraybackslash}m{0.28\textwidth}
}
\toprule

\textbf{Task Setup} &
\textbf{Description} &
\textbf{Task Setup} &
\textbf{Description}
\\

\midrule

\includegraphics[width=0.19\textwidth]{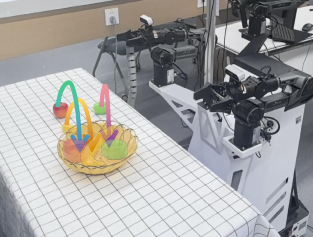}
&
\textbf{Pick Fruits:}
Pick the fruits and put them in the basket.

&
\includegraphics[width=0.19\textwidth]{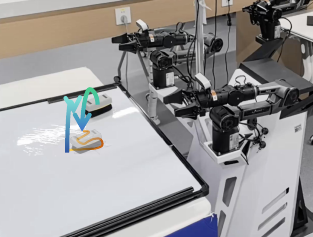}
&
\textbf{Wipe Board:}
Pick up the eraser and wipe the whiteboard.
\\

\midrule

\includegraphics[width=0.19\textwidth]{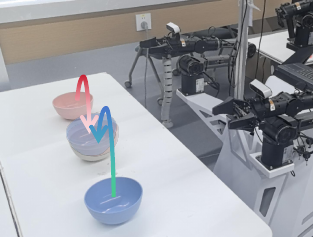}
&
\textbf{Stack Bowls:}
Stack the bowls on the table, with the middle one.

&
\includegraphics[width=0.19\textwidth]{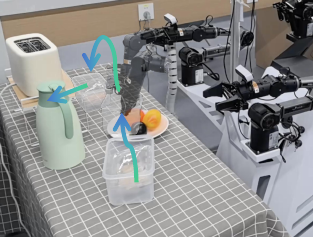}
&
\textbf{Box Return:}
Move the box under the shelf.
\\

\midrule

\includegraphics[width=0.19\textwidth]{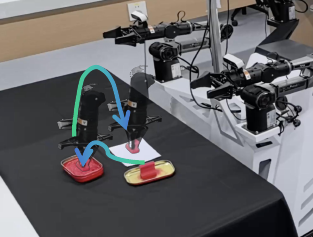}
&
\textbf{Seal Stamping:}
Dip the stamp in ink and stamp.

&
\includegraphics[width=0.19\textwidth]{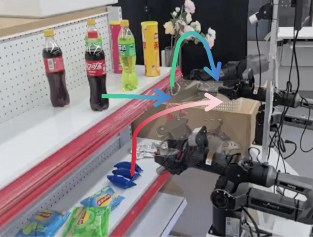}
&
\textbf{Shelf Retrieval:}
Use the left arm to retrieve a Coke from the upper shelf and transfer it to the right arm for placement into the basket, then use the right arm to retrieve Oreos from the lower shelf and place them into the basket.
\\

\bottomrule
\end{tabular}
\end{table*}

We summarize the six real-robot tasks in Table~\ref{tab:appendix_task_descriptions}, including task names, execution illustrations, and policy prompts.

All real-robot experiments are conducted on an AgileX Cobot Magic platform, a Mobile ALOHA-style bimanual robot with two forward-reaching arms. The robot uses three RGB cameras: one upper-body camera and two wrist cameras mounted near the left and right end-effectors. During data collection, we record synchronized RGB observations from the three cameras and teleoperated expert trajectories. We fully fine-tune the $\pi_0$, a VLA model, using the OpenPI JAX framework on the collected real-robot demonstrations, with a batch size of 32 for 50k steps. During deployment, an onboard NVIDIA RTX 4090 GPU is used for real-time action inference and execution.

\subsection{Baselines}
\label{app:B.3}

We compare \textsc{ATHENA} to several data curation baselines, including Oracle, Random, TAROT, TSS, and Distillation. For all baselines, we use the same filtering ratios
$\rho\in\{0.10,0.25,0.50,0.75,0.90\}$. All curated subsets are fine-tuned and evaluated with the same settings as \textsc{ATHENA}.

\paragraph{\textbf{Oracle.}}
Following the Oracle definitions in RoboMimic and CUPID~\cite{mandlekar2021robomimic,agia2025cupid}, demonstration length is used as a proxy for dataset quality. Specifically, Oracle assumes that shorter length demonstrations correspond to more efficient task completion and therefore indicate higher demonstration quality. For each demonstration $\xi_i$, the Oracle baseline assigns the following quality score:
\begin{equation}
    S_{\mathrm{oracle}}(\xi_i) = -T(\xi_i),
\end{equation}
where $T(\xi_i)$ denotes the demonstration length, i.e., the completion time of $\xi_i$. Demonstrations are ranked in descending order of $S_{\mathrm{oracle}}$, so demonstrations with shorter length are retained before longer ones. However, Oracle does not accurately model downstream task success, and as shown in Figures~\ref{fig:quality_only_50task} and ~\ref{fig:multitask_quality_success_50} (see also CUPID~\cite{agia2025cupid}), the expert-based quality scores do not always align with downstream performance.

\paragraph{\textbf{Random.}}
Random is a data-agnostic baseline that uniformly samples demonstrations from the training pool under each filtering ratio. 

\paragraph{\textbf{TAROT.}}

We adapt TAROT~\cite{pmlr-v267-feng25l} as a target-distribution matching baseline for multitask demonstration curation. It selects training demonstrations by matching candidate demonstrations to a target set in a whitened gradient-feature space. In our setting, the candidate set consists of training demonstrations, while the target set consists of held-out or evaluation episodes from the corresponding task suite. TAROT constructs a candidate-target similarity matrix and solves an entropic optimal-transport problem with Sinkhorn iterations. The transport plan prioritizes demonstrations that better cover the target distribution. However, unlike \textsc{ATHENA}, TAROT remains sensitive to target-set quality, limiting its multitask curation performance.

\paragraph{\textbf{Temporal Surprise Score (TSS).}}
TSS~\cite{yang2026less}  is a temporal-difficulty sampling baseline that values demonstrations by the magnitude of their action changes over time. 
For RoboTwin 2.0 ALOHA tasks, each action $a_t$ is a $14$-dimensional dual-arm command, consisting of two $6$-D arm-motion commands and two scalar gripper commands. 
We denote the arm-motion component as
$m_t=[a_t^0,\ldots,a_t^5,a_t^7,\ldots,a_t^{12}]$ and the two gripper commands as $g_t^L=a_t^6$ and $g_t^R=a_t^{13}$. 
The frame-level TSS score is defined as
\begin{equation}
    S(t)
    =
    D_{\cos}(m_t,m_{t-1})
    +
    \gamma
    \left(
    |g_t^L-g_{t-1}^L|
    +
    |g_t^R-g_{t-1}^R|
    \right),
\end{equation}
where $D_{\cos}$ is the cosine distance between consecutive arm-motion commands, and $\gamma$ controls the relative weight of gripper changes. 
We set $\gamma=1$ in all experiments. 
We aggregate frame-level scores into a demonstration-level score and rank demonstrations in descending order, so trajectories with larger temporal action changes are retained first.

\paragraph{\textbf{Distillation.}}
Distillation is a demonstration-level baseline that selects trajectories based on their diversity with respect to task-level action patterns. 
For each task, we compute a prototype trajectory from the average resampled demonstrations, and score each demonstration by its deviation from this prototype:
\begin{equation}
    S_{\mathrm{distill}}(\xi_i)
    =
    \left\|
    \mathrm{Resample}(\xi_i)
    -
    \mu_{\mathrm{task}(i)}
    \right\|_{F},
\end{equation}
where $\mu_{\mathrm{task}(i)}$ denotes the task-level prototype trajectory, computed by averaging the resampled demonstrations from the same task as $\xi_i$. 
Demonstrations with larger scores are treated as less redundant and are retained first, encouraging diversity in the curated subset. 
However, this score is only a trajectory-space proxy: it does not use policy rollouts, gradient information, or downstream performance feedback, and may therefore select diverse demonstrations that are not necessarily beneficial for improving policy success.

\section{Additional Experiments and Discussion}
\label{app:C}

\subsection{Analysis of Computational Efficiency}
\label{app:C.1}

We ablate ATHENA's two core acceleration components, Kronecker compressed gradient featurization and the Random Truncated Approximation for the Hessian, to demonstrate their necessity for scaling influence calculation to large-scale, widely adopted VLA models. We conduct this evaluation on a multitask setup consisting of $K=50$ tasks with $N \approx 560{,}500$ total timesteps, benchmarking the 3.3B-parameter $\pi_0$ model as a representative case on a 140\,GB GPU platform (Table~\ref{tab:logra_ablation}).

As shown, directly applying naive CUPID to $\pi_0$ is computationally prohibitive. Explicitly materializing the full $D$-dimensional gradient causes immediate out-of-memory errors even on high-end 140\,GB hardware, forcing a strict batch size of 1 and inflating featurization time to 8004.6\,GPU$\cdot$h. ATHENA circumvents this spatial bottleneck via Kronecker compressed projection directly onto layer-local activation buffers, reducing peak memory to $\sim 1$\,GB and accelerating featurization to 23.2\,GPU$\cdot$h under a distributed batched setting. For the Hessian step, dense inversion requires 50\,GPU$\cdot$h, whereas ATHENA's truncation approximation requires only 2.5\,GPU$\cdot$h. Combined, ATHENA reduces the total 50-task attribution overhead from 8054.6 to 25.7\,GPU$\cdot$h, achieving a $313.4\times$ speedup. Crucially, while a cost of nearly 8054.6\,GPU$\cdot$h on a modest 50-task setup fundamentally shuts down any possibility of scaling naive methods to broader multitask regimes, ATHENA's high efficiency unlocks a viable path toward open-ended scaling across hundreds of tasks and millions of timesteps.

\begin{table}[h]
    \centering
    \footnotesize 
    \setlength{\tabcolsep}{4pt}

    \caption{\textbf{Complexity ablation on $\pi_0$ for $K{=}50$ tasks.} VRAM is incremental, and GPU$\cdot$h influences computation time. $^\dagger$ marks per-sample full-gradient memory.}
    \label{tab:logra_ablation}

    \begin{tabular}{lccc}
        \toprule
         & Time complexity
         & Incremental VRAM
         & GPU$\cdot$h                                      \\
        \midrule
        \multicolumn{4}{l}{\itshape Gradient featurization (50 tasks)} \\[2pt]
        w/o Kronecker
         & $\mathcal{O}(KN \cdot DP)$
         & $\sim$13\,GB$^\dagger$ (+OOM)
         & 8004.6                                           \\
        ATHENA
         & $\mathcal{O}(KN \cdot Lkd)$
         & $\mathbf{\sim 1}$\,GB
         & $\mathbf{23.2}$                                  \\
        \midrule
        \multicolumn{4}{l}{\itshape Hessian approximation}             \\[2pt]
        w/o RTA
         & $\mathcal{O}(NP^2 + P^3)$
         & —
         & 50.0                                             \\
        ATHENA (Random Truncation)
         & $\mathcal{O}(KN \cdot D_{\mathrm{grad}}r)$
         & —
         & $\mathbf{2.5}$                                   \\
        \midrule
        \textbf{Total, 50 tasks}
         & —
         & —
         & 8054.6 \textbf{vs} 25.7                          \\
        \bottomrule
    \end{tabular}
\end{table}

\subsection{Per-Setting Multitask Ablation Results}
\label{app:multitask_results}

\begin{figure}[h]
    \centering
    \includegraphics[width=\linewidth]{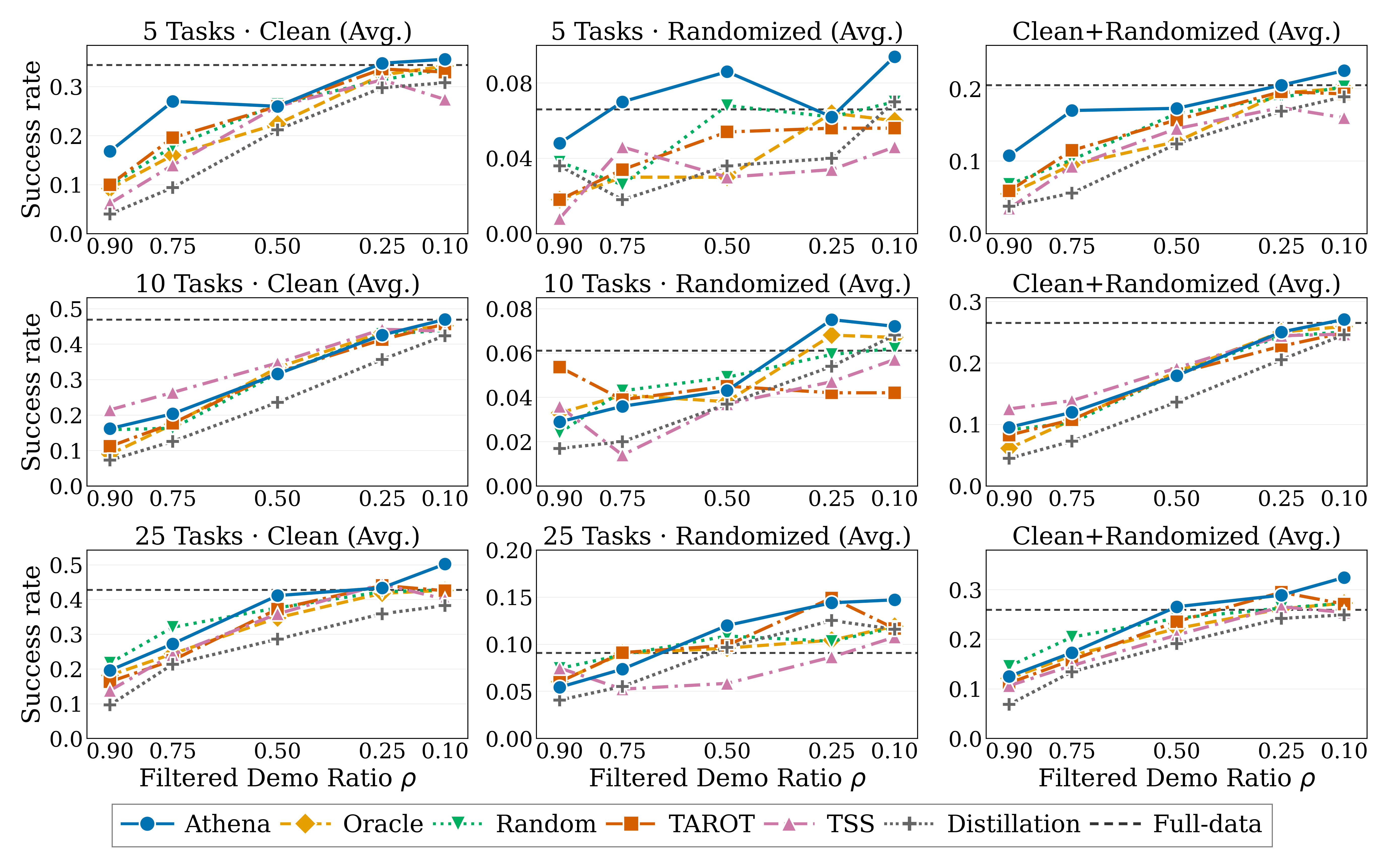}
    \vspace{-0.6cm}
    \caption{
        Success rate trends across task counts and filtering ratios $\rho$.
        Each row corresponds to $K\in\{5,10,25\}$, and columns report clean,
        randomized, and averaged success rates. Dashed lines denote full-data fine-tuning.
    }
    \label{fig:multitask_ablation_success_5_10_25}
\end{figure}
This section complements Section~\ref{sec:6.2} by decomposing the averaged multitask results into clean and randomized evaluations. As shown in
Figure~\ref{fig:multitask_ablation_success_5_10_25}, ATHENA remains competitive
across both settings and all task scales, indicating that the gains reported in the main text are not dominated by a single evaluation setting.

At $\rho=0.1$, ATHENA exceeds full-data performance under both evaluations for
all three task scales. For $K=5$, it achieves {35.60\%} clean-evaluation success
and {9.40\%} randomized-evaluation success, compared with full-data baselines
of {34.40\%} and {6.60\%}. For $K=10$, ATHENA reaches {47.00\%} and {7.21\%},
surpassing {46.90\%} and {6.11\%}. For $K=25$, the gains become more pronounced,
with {50.24\%} clean-evaluation success and {14.72\%} randomized-evaluation
success, compared with {42.76\%} and {9.09\%}.

At $\rho=0.5$, the per-setting curves are consistent with the data-scale effect
discussed in Section~6.2. Smaller task subsets ($K=5,10$) provide a limited
candidate set for curation, leading to weaker or unstable gains, whereas
$K=25$ allows ATHENA to remove low-utility demonstrations while preserving task
coverage. These results provide per-setting evidence for the main-text observation that curation yields higher returns as task diversity increases.

\subsection{Per-task Success Analysis}
\label{app:appendix_per_task}

Tables~\ref{tab:per_task_success_clean_randomized_full} and~\ref{tab:pi0_selected_pi05_per_task_success} report per-task evaluation results of jointly fine-tuned 50-task policies under different filtering ratios. 
We refer to Lingbo-VA~\cite{lingbo-va} to stratify the 50 tasks into 30 H1, 15 H2, and 5 H3.

For the detailed breakdown of Figure~\ref{fig:multitask_quality_success_50}, corresponding to the main $\pi_0\!\rightarrow\!\pi_0$ setting (Table~\ref{tab:per_task_success_clean_randomized_full}), at the aggregate level $\rho=0.5$ preserves clean performance ($43.42\% \rightarrow 43.36\%$) while improving randomized success from $15.44\%$ to $17.30\%$. 
With lighter filtering ($\rho=0.1$), both clean and randomized averages increase to $44.70\%$ and $17.72\%$, respectively. 
Specifically, per-horizon analysis highlights contributions from different task groups.

For H1 tasks, $\rho=0.1$ increases the clean average from $47.7\%$ to $50.1\%$ and randomized from $19.5\%$ to $21.8\%$, with notable gains on contact- or alignment-sensitive tasks, e.g., \textit{place\_shoe} ($43.0\%\!\rightarrow\!65.0\%$), \textit{click\_bell} ($66.0\%\!\rightarrow\!83.0\%$), and \textit{press\_stapler} ($70.0\%\!\rightarrow\!75.0\%$). 
For H2 tasks, moderate filtering ($\rho=0.5$) yields the best group averages: clean from $39.1\%$ to $40.8\%$, randomized from $10.47\%$ to $13.27\%$, with representative gains on \textit{place\_cans\_plasticbox} ($55.0\%\!\rightarrow\!77.0\%$) and \textit{place\_bread\_skillet} ($35.0\%\!\rightarrow\!49.0\%$). 
H3 tasks benefit mainly in clean evaluation at $\rho=0.1$ ($24.8\% \rightarrow 28.0\%$) and in randomized at $\rho=0.5$ ($5.8\% \rightarrow 9.6\%$), e.g., \textit{stack\_bowls\_three} ($61.0\%\!\rightarrow\!72.0\%$) and \textit{blocks\_ranking\_rgb} ($16.0\%\!\rightarrow\!20.0\%$). 
These observations demonstrate that highly redundant tasks benefit from removing low-utility data, whereas precise or long-horizon tasks are more sensitive and may see declines when essential data is removed~\cite{agia2025cupid}.

Table~\ref{tab:pi0_selected_pi05_per_task_success} shows that the same $\pi_0$-curated subsets also improve $\pi_{0.5}$ performance (corresponding to Table~\ref{tab:pi0_guides_pi05}). Using these subsets increases the average to $67.30\%$ clean and $37.77\%$ randomized at $\rho=0.1$, and even with only half of the data retained ($\rho=0.5$), it reaches $63.28\%$ clean and $34.23\%$ randomized, both exceeding the full-data baselines ($57.00\%$ clean, $25.68\%$ randomized), demonstrating cross-model applicability.

\subsection{Retention Balance, Single-Task Curation, and Real-Robot Failure Modes}
\label{app:MII}
\begin{figure*}[t]
    \centering
    \includegraphics[width=\textwidth]{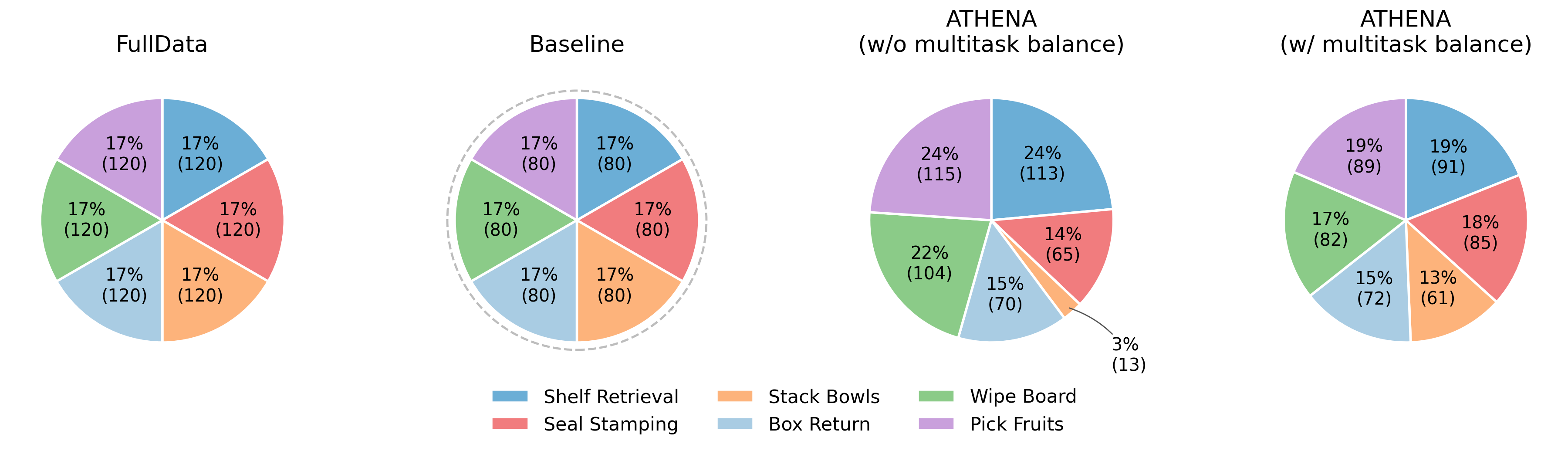}
\caption{Distributions of the retained 66.7\% demonstrations across the six real-robot tasks.}
    \label{fig:athena_multitask_balance}
\end{figure*}

To further ablate the role of Multitask Influence Interaction (MII), we visualize the retained task distributions after data curation in Fig.~\ref{fig:athena_multitask_balance}. 
We consider the six-task real-robot setting with 120 demonstrations per task and an overall retention ratio of 66.7\%. 
Without MII, naively ranking demonstrations with a single global influence score results in a highly skewed retained set: Pick Fruits, Shelf Retrieval, and Wipe Board retain 115, 113, and 104 demonstrations, respectively, whereas Stack Bowls retains only 13 and is nearly eliminated. 
This occurs because raw influence magnitudes are not directly comparable across heterogeneous tasks with different horizons, motion patterns, and cross-task coupling; as a result, the naive global ranking can overemphasize dataset-level contribution while ignoring task-local importance. 
With MII, ATHENA combines task-local and cross-task influence utilities, yielding a balanced retained distribution and preventing task-level collapse under the same retention ratio.

\paragraph{\textit{RoboTwin} single-task analysis.}

\begin{figure}[t] 
    \centering 
    \includegraphics[width=\linewidth]{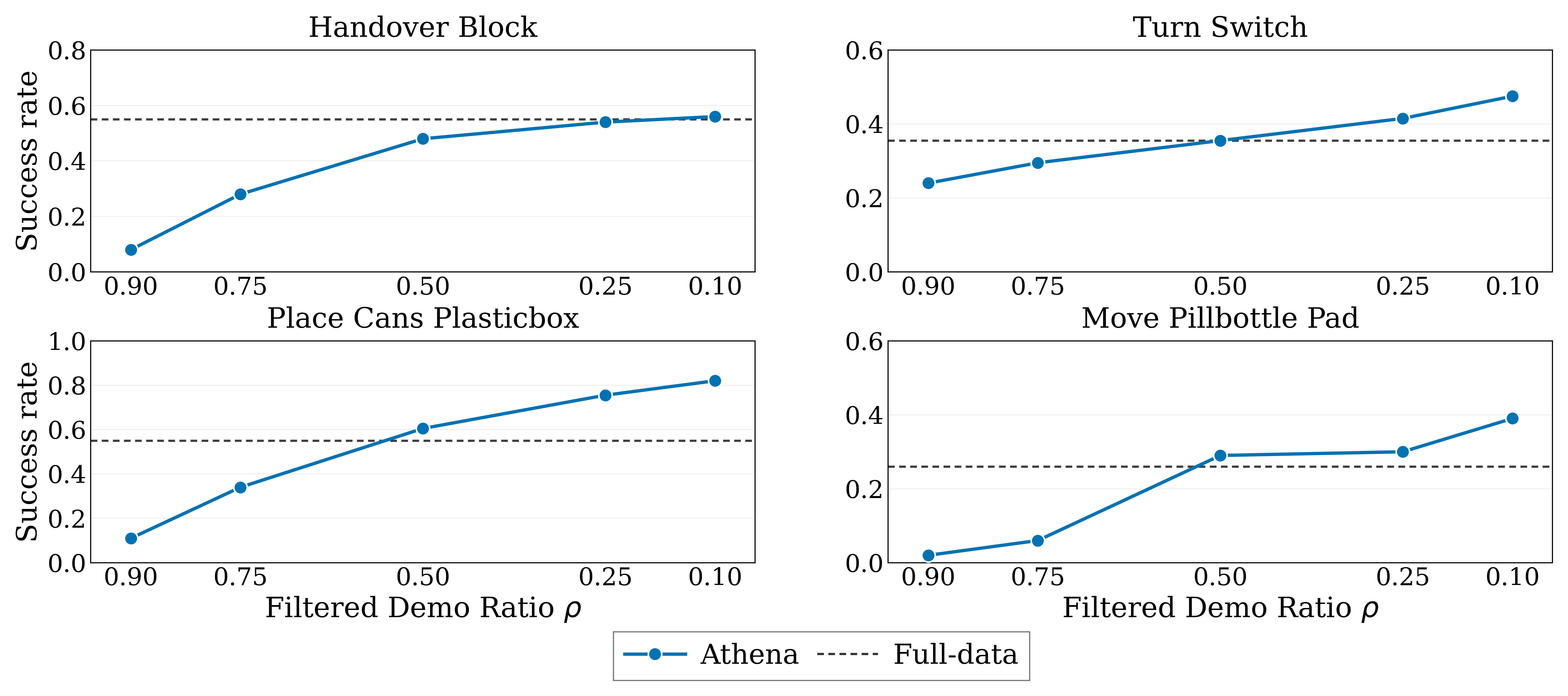} \caption{Single-task fine-tuning results on four RoboTwin tasks. } 
    \label{fig:single_task_if} 
\end{figure}

Fig.~\ref{fig:single_task_if} reports single-task fine-tuning results on four RoboTwin tasks. Light curation consistently improves policy performance: filtering only $10\%$ of demonstrations outperforms full-data training on all tasks, improving success from $55\%$ to $56\%$ on handover, $35.5\%$ to $47.5\%$ on turn switch, $55\%$ to $82\%$ on place cans plasticbox, and $26\%$ to $39\%$ on move pillbottle pad. This suggests that a fraction of demonstrations can harm policy performance rather than improve it, consistent with prior findings in influence-based robot data curation~\cite{agia2025cupid}. With stronger filtering, performance becomes task-dependent: handover is more data-hungry and degrades as more demonstrations are removed, likely because it requires sufficient coverage of diverse interaction and transfer states; in contrast, the other three tasks remain robust to data reduction. At $\rho=0.5$, \textsc{ATHENA} matches or surpasses full-data training on three out of four tasks.

\begin{figure*}[t]
    \centering
    \includegraphics[width=\textwidth]{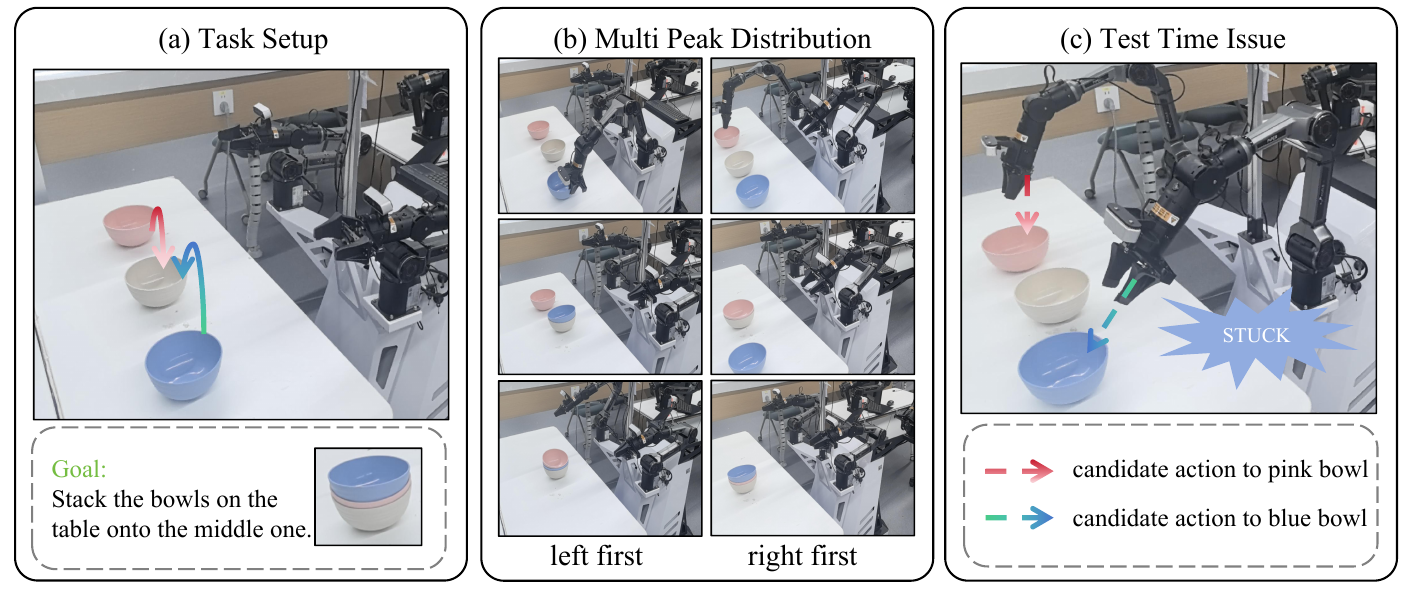}
    \caption{
    \textbf{Illustration of the multi-peak action distribution in the Stack Bowls task.} (a) The task requires stacking the bowls onto the middle bowl. (b) The collected demonstrations contain two valid bimanual execution modes, corresponding to left-first and right-first manipulation orders. (c) At test time, the learned policy may mix the two modes and command the two arms toward different candidate bowls simultaneously, leading to execution hesitation or a stuck state.
    }
    \label{fig:stack_bowls_multipeak}
\end{figure*}

\paragraph{\textit{Multi-peak Action Distribution} task analysis.}
As shown in Fig.~\ref{fig:stack_bowls_multipeak}, the Stack Bowls task exhibits a multi-peak action distribution in the collected demonstrations. Specifically, under similar task setups, the demonstrations contain two valid bimanual execution modes: a left-first mode accounting for roughly one-third of the demonstrations, where the left arm initiates the stacking sequence before the right arm, and a right-first mode accounting for roughly two-thirds, where the execution order is reversed. Although both modes can complete the task, their coexistence in the training data introduces ambiguity into the learned action distribution. During inference, the policy may combine action tendencies from both modes, causing the two arms to move toward different candidate bowls simultaneously and leading to execution hesitation or occasional stalling. Applying ATHENA to curate the task-specific training data increases the success rate from 56\% to 68\%, suggesting that data curation mitigates the effect of multi-peak action modes. However, its success rate remains limited when evaluated across different bowl positions, as the policy lacks compositional generalization over spatial bowl arrangements without multitask training.

\begin{figure*}[t]
    \centering
    \includegraphics[width=\textwidth]{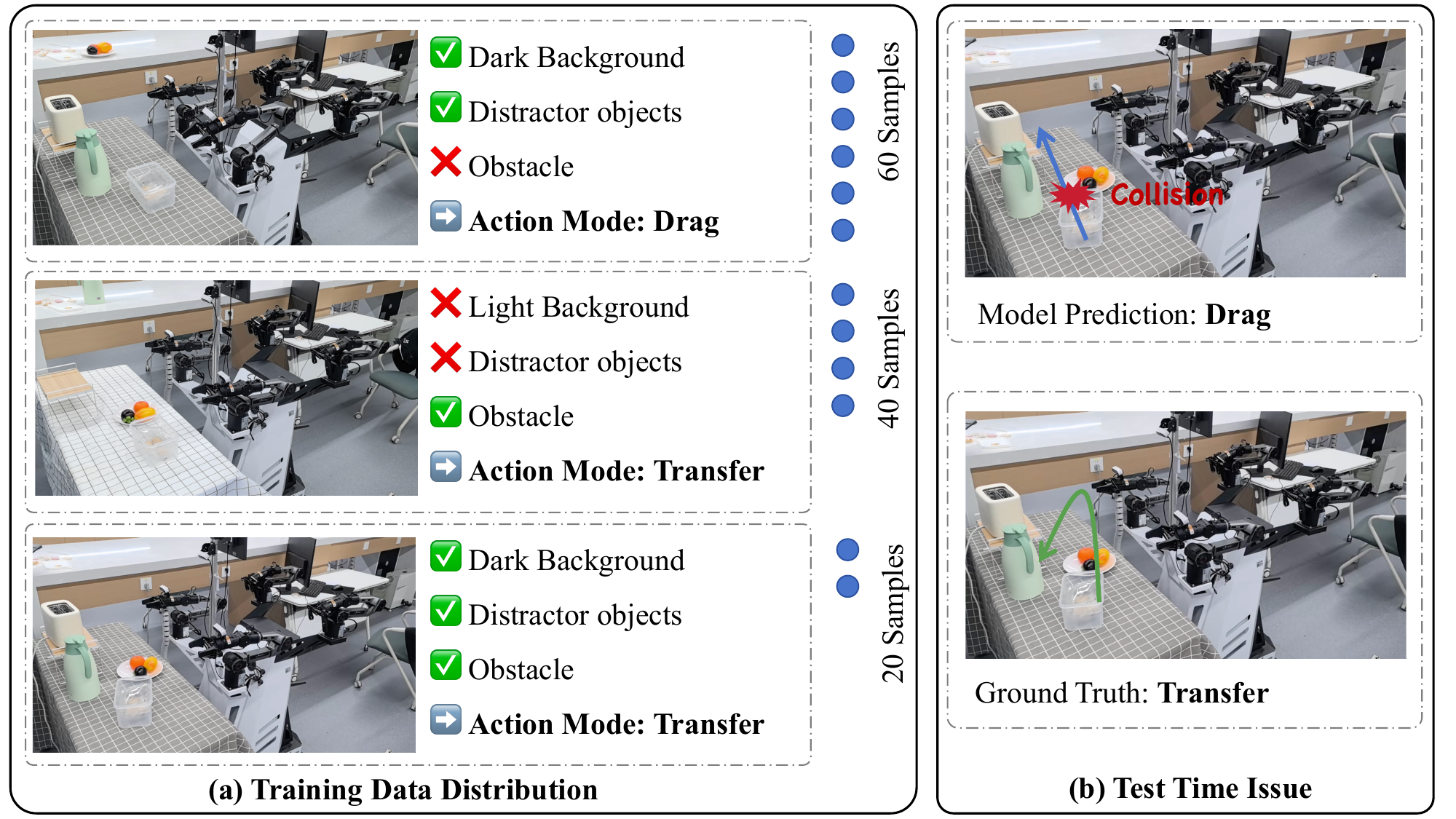}
    \caption{
    \textbf{Illustration of spurious associations in the Box Return task.}
    (a) The collected demonstrations contain imbalanced subgroups that couple visual appearance, obstacle presence, and action modes.
    The smallest subgroup provides disambiguating examples where distractor objects and an obstacle coexist, but the correct behavior remains transfer.
    (b) At test time, the policy may rely on spurious visual cues and predict dragging when transfer is required, leading to obstacle collision.
    }
  
    \label{fig:box_return_spurious}
\end{figure*}
\paragraph{\textit{Spurious Association} task analysis.}
As shown in Fig.~\ref{fig:box_return_spurious}, the Box Return task exhibits a spurious association between scene appearance and action mode in the collected demonstrations. The training data contains three subgroups with an approximate ratio of 3:2:1. In the largest subgroup, the scene uses a dark tablecloth with distractor objects and no obstacle, and the demonstrated behavior follows a dragging mode. In the second subgroup, the scene uses a light tablecloth with no distractor objects but includes an obstacle, and the demonstrated behavior follows a transfer mode. The smallest subgroup contains both distractor objects and an obstacle, while still requiring the transfer mode. Due to the imbalance among these subgroups, the policy can incorrectly associate background appearance or distractor objects with the dragging behavior, rather than relying on the obstacle configuration to determine the appropriate action mode. At test time, this spurious association may cause the policy to predict dragging even when a transfer is required, leading to a collision with the obstacle. By applying ATHENA to curate the task-specific training data, the relative contribution of the disambiguating subgroup is increased, reducing the effect of spurious visual correlations and improving the selection of the correct action mode. This leads to a substantial improvement in performance, increasing the success rate from 52\% to 74\%.

\begin{table*}[p]
\centering
\caption{Per-task success rates under different filtering ratios on the 50-task benchmark. Values are success rates in percentage points. The full-data result is denoted by $\rho=0.0$.}
\label{tab:per_task_success_clean_randomized_full}
\setlength{\tabcolsep}{2.4pt}
\renewcommand{\arraystretch}{1.0}
\resizebox{\textwidth}{!}{%
\begin{tabular}{lc*{12}{c}}
\toprule
Task & Horizon & \multicolumn{2}{c}{$\rho=0.9$} & \multicolumn{2}{c}{$\rho=0.75$} & \multicolumn{2}{c}{$\rho=0.5$} & \multicolumn{2}{c}{$\rho=0.25$} & \multicolumn{2}{c}{$\rho=0.1$} & \multicolumn{2}{c}{$\rho=0.0$} \\
\cmidrule(lr){3-4} \cmidrule(lr){5-6} \cmidrule(lr){7-8} \cmidrule(lr){9-10} \cmidrule(lr){11-12} \cmidrule(lr){13-14}
 & & Clean & Rand. & Clean & Rand. & Clean & Rand. & Clean & Rand. & Clean & Rand. & Clean & Rand. \\
\midrule
Adjust Bottle & 1 & 35.00 & 7.00 & 65.00 & 3.00 & 62.00 & 4.00 & 41.00 & 11.00 & 56.00 & 9.00 & 70.00 & 10.00 \\
Beat Block Hammer & 1 & 17.00 & 7.00 & 28.00 & 12.00 & 62.00 & 12.00 & 62.00 & 11.00 & 64.00 & 13.00 & 60.00 & 13.00 \\
Blocks Ranking Rgb & 3 & 0.00 & 0.00 & 5.00 & 0.00 & 14.00 & 2.00 & 45.00 & 2.00 & 20.00 & 1.00 & 16.00 & 3.00 \\
Blocks Ranking Size & 3 & 0.00 & 0.00 & 3.00 & 0.00 & 8.00 & 1.00 & 0.00 & 0.00 & 9.00 & 0.00 & 11.00 & 1.00 \\
Click Alarmclock & 1 & 78.00 & 33.00 & 82.00 & 27.00 & 78.00 & 41.00 & 90.00 & 32.00 & 86.00 & 47.00 & 85.00 & 19.00 \\
Click Bell & 1 & 91.00 & 54.00 & 92.00 & 39.00 & 90.00 & 49.00 & 83.00 & 39.00 & 83.00 & 45.00 & 66.00 & 18.00 \\
Dump Bin Bigbin & 1 & 42.00 & 30.00 & 55.00 & 49.00 & 74.00 & 34.00 & 69.00 & 37.00 & 63.00 & 30.00 & 75.00 & 37.00 \\
Grab Roller & 1 & 64.00 & 41.00 & 79.00 & 48.00 & 93.00 & 60.00 & 100.00 & 52.00 & 93.00 & 75.00 & 99.00 & 67.00 \\
Handover Block & 2 & 0.00 & 0.00 & 11.00 & 1.00 & 15.00 & 0.00 & 11.00 & 0.00 & 21.00 & 0.00 & 5.00 & 0.00 \\
Handover Mic & 2 & 61.00 & 1.00 & 57.00 & 2.00 & 29.00 & 2.00 & 51.00 & 2.00 & 35.00 & 6.00 & 49.00 & 4.00 \\
Hanging Mug & 2 & 2.00 & 0.00 & 10.00 & 3.00 & 10.00 & 1.00 & 6.00 & 2.00 & 15.00 & 2.00 & 9.00 & 0.00 \\
Lift Pot & 1 & 8.00 & 0.00 & 16.00 & 0.00 & 23.00 & 0.00 & 26.00 & 0.00 & 29.00 & 0.00 & 20.00 & 0.00 \\
Move Can Pot & 1 & 5.00 & 0.00 & 12.00 & 0.00 & 48.00 & 7.00 & 40.00 & 2.00 & 10.00 & 2.00 & 19.00 & 1.00 \\
Move Pillbottle Pad & 1 & 4.00 & 0.00 & 14.00 & 1.00 & 19.00 & 3.00 & 21.00 & 2.00 & 24.00 & 7.00 & 30.00 & 5.00 \\
Move Playingcard Away & 1 & 20.00 & 1.00 & 61.00 & 19.00 & 71.00 & 32.00 & 59.00 & 15.00 & 71.00 & 17.00 & 56.00 & 30.00 \\
Move Stapler Pad & 1 & 0.00 & 0.00 & 2.00 & 0.00 & 1.00 & 1.00 & 1.00 & 0.00 & 1.00 & 0.00 & 1.00 & 0.00 \\
Open Laptop & 1 & 55.00 & 23.00 & 78.00 & 19.00 & 67.00 & 20.00 & 66.00 & 12.00 & 58.00 & 13.00 & 49.00 & 21.00 \\
Open Microwave & 1 & 14.00 & 24.00 & 34.00 & 29.00 & 57.00 & 16.00 & 53.00 & 25.00 & 41.00 & 45.00 & 60.00 & 39.00 \\
Pick Diverse Bottles & 2 & 6.00 & 1.01 & 48.00 & 7.07 & 50.00 & 3.03 & 47.00 & 1.01 & 59.00 & 5.05 & 48.00 & 2.02 \\
Pick Dual Bottles & 2 & 5.00 & 1.00 & 55.00 & 10.00 & 67.00 & 12.00 & 56.00 & 10.00 & 62.00 & 18.00 & 79.00 & 7.00 \\
Place A2b Left & 1 & 15.00 & 1.00 & 19.00 & 2.00 & 23.00 & 11.00 & 24.00 & 9.00 & 19.00 & 8.00 & 24.00 & 7.00 \\
Place A2b Right & 1 & 8.00 & 3.00 & 8.00 & 4.00 & 19.00 & 4.00 & 25.00 & 4.00 & 27.00 & 6.00 & 31.00 & 13.00 \\
Place Bread Basket & 1 & 4.00 & 1.00 & 14.00 & 4.00 & 31.00 & 20.00 & 30.00 & 27.00 & 37.00 & 18.00 & 38.00 & 31.00 \\
Place Bread Skillet & 2 & 1.00 & 0.00 & 17.00 & 2.00 & 49.00 & 11.00 & 61.00 & 13.00 & 33.00 & 17.00 & 35.00 & 17.00 \\
Place Burger Fries & 2 & 9.00 & 4.00 & 5.00 & 1.00 & 39.00 & 58.00 & 11.00 & 32.00 & 9.00 & 34.00 & 17.00 & 36.00 \\
Place Can Basket & 2 & 18.00 & 0.00 & 28.00 & 1.00 & 51.00 & 8.00 & 27.00 & 4.00 & 47.00 & 2.00 & 46.00 & 1.00 \\
Place Cans Plasticbox & 2 & 18.00 & 6.00 & 17.00 & 11.00 & 77.00 & 16.00 & 16.00 & 28.00 & 51.00 & 28.00 & 55.00 & 37.00 \\
Place Container Plate & 1 & 69.00 & 43.00 & 72.00 & 44.00 & 35.00 & 36.00 & 40.00 & 23.00 & 64.00 & 51.00 & 52.00 & 23.00 \\
Place Dual Shoes & 2 & 4.00 & 6.00 & 18.00 & 10.00 & 42.00 & 12.00 & 34.00 & 14.00 & 41.00 & 31.00 & 43.00 & 19.00 \\
Place Empty Cup & 1 & 24.00 & 7.00 & 27.00 & 7.00 & 52.00 & 18.00 & 64.00 & 14.00 & 81.00 & 30.00 & 72.00 & 35.00 \\
Place Fan & 1 & 2.00 & 1.00 & 4.00 & 2.00 & 4.00 & 0.00 & 7.00 & 2.00 & 20.00 & 5.00 & 17.00 & 7.00 \\
Place Mouse Pad & 1 & 4.00 & 0.00 & 15.00 & 4.00 & 15.00 & 2.00 & 28.00 & 5.00 & 20.00 & 1.00 & 13.00 & 4.00 \\
Place Object Basket & 2 & 24.00 & 2.00 & 58.00 & 7.00 & 68.00 & 22.00 & 62.00 & 7.00 & 71.00 & 3.00 & 72.00 & 3.00 \\
Place Object Scale & 1 & 5.00 & 0.00 & 14.00 & 3.00 & 16.00 & 6.00 & 20.00 & 3.00 & 22.00 & 2.00 & 9.00 & 3.00 \\
Place Object Stand & 1 & 35.00 & 19.00 & 45.00 & 16.00 & 68.00 & 28.00 & 47.00 & 19.00 & 57.00 & 27.00 & 41.00 & 7.00 \\
Place Phone Stand & 1 & 8.00 & 0.00 & 14.00 & 3.00 & 25.00 & 7.00 & 34.00 & 2.00 & 32.00 & 2.00 & 35.00 & 5.00 \\
Place Shoe & 1 & 5.00 & 3.00 & 34.00 & 9.00 & 45.00 & 26.00 & 40.00 & 22.00 & 65.00 & 33.00 & 43.00 & 18.00 \\
Press Stapler & 1 & 62.00 & 15.00 & 55.00 & 20.00 & 57.00 & 24.00 & 69.00 & 22.00 & 75.00 & 16.00 & 70.00 & 18.00 \\
Put Bottles Dustbin & 3 & 14.00 & 1.00 & 35.00 & 8.00 & 25.00 & 6.00 & 26.00 & 6.00 & 37.00 & 5.00 & 31.00 & 8.00 \\
Put Object Cabinet & 2 & 18.00 & 1.00 & 35.00 & 1.00 & 24.00 & 3.00 & 18.00 & 1.00 & 17.00 & 2.00 & 19.00 & 2.00 \\
Rotate QRcode & 1 & 39.00 & 3.00 & 53.00 & 12.00 & 54.00 & 12.00 & 57.00 & 2.00 & 63.00 & 5.00 & 66.00 & 8.00 \\
Scan Object & 2 & 5.00 & 1.00 & 11.00 & 1.00 & 11.00 & 2.00 & 11.00 & 4.00 & 12.00 & 7.00 & 13.00 & 6.00 \\
Shake Bottle & 1 & 87.00 & 56.00 & 97.00 & 54.00 & 92.00 & 71.00 & 92.00 & 67.00 & 96.00 & 65.00 & 98.00 & 68.00 \\
Shake Bottle Horizontally & 1 & 89.00 & 52.00 & 96.00 & 53.00 & 90.00 & 60.00 & 93.00 & 64.00 & 94.00 & 56.00 & 90.00 & 68.00 \\
Stack Blocks Three & 3 & 0.00 & 0.00 & 0.00 & 0.00 & 2.00 & 0.00 & 2.00 & 1.00 & 2.00 & 0.00 & 5.00 & 0.00 \\
Stack Blocks Two & 2 & 14.00 & 2.00 & 8.00 & 2.00 & 23.00 & 1.00 & 27.00 & 13.00 & 26.00 & 3.00 & 37.00 & 5.00 \\
Stack Bowls Three & 3 & 28.00 & 9.00 & 55.00 & 20.00 & 62.00 & 39.00 & 66.00 & 25.00 & 72.00 & 33.00 & 61.00 & 17.00 \\
Stack Bowls Two & 2 & 61.00 & 18.00 & 87.00 & 33.00 & 87.00 & 48.00 & 90.00 & 38.00 & 92.00 & 35.00 & 89.00 & 18.00 \\
Stamp Seal & 1 & 10.00 & 0.00 & 10.00 & 3.00 & 20.00 & 1.00 & 18.00 & 6.00 & 22.00 & 8.00 & 16.00 & 2.00 \\
Turn Switch & 1 & 10.00 & 8.00 & 11.00 & 14.00 & 24.00 & 13.00 & 34.00 & 10.00 & 31.00 & 18.00 & 26.00 & 9.00 \\
\midrule
\textbf{Average} & -- & \textbf{23.94} & \textbf{9.70} & \textbf{35.38} & \textbf{12.40} & \textbf{43.36} & \textbf{17.30} & \textbf{42.00} & \textbf{14.84} & \textbf{44.70} & \textbf{17.72} & \textbf{43.42} & \textbf{15.44} \\
\bottomrule
\end{tabular}%
}
\end{table*}

\begin{table*}[p]
\centering
\caption{Per-task success rates when using $\pi_0$-selected data to train $\pi_{0.5}$. Values are success rates in percentage points. The full-data result is denoted by $\rho=0.0$.}
\label{tab:pi0_selected_pi05_per_task_success}
\setlength{\tabcolsep}{2.4pt}
\renewcommand{\arraystretch}{1.0}
\resizebox{\textwidth}{!}{%
\begin{tabular}{lc*{12}{c}}
\toprule
Task & Horizon & \multicolumn{2}{c}{$\rho=0.9$} & \multicolumn{2}{c}{$\rho=0.75$} & \multicolumn{2}{c}{$\rho=0.5$} & \multicolumn{2}{c}{$\rho=0.25$} & \multicolumn{2}{c}{$\rho=0.1$} & \multicolumn{2}{c}{$\rho=0.0$} \\
\cmidrule(lr){3-4} \cmidrule(lr){5-6} \cmidrule(lr){7-8} \cmidrule(lr){9-10} \cmidrule(lr){11-12} \cmidrule(lr){13-14}
 & & Clean & Rand. & Clean & Rand. & Clean & Rand. & Clean & Rand. & Clean & Rand. & Clean & Rand. \\
\midrule
Adjust Bottle & 1 & 61.00 & 39.00 & 44.00 & 48.00 & 97.00 & 75.00 & 100.00 & 89.00 & 100.00 & 87.00 & 97.00 & 54.00 \\
Beat Block Hammer & 1 & 56.00 & 8.00 & 71.00 & 7.00 & 78.00 & 10.00 & 90.00 & 32.00 & 80.00 & 14.00 & 3.00 & 0.00 \\
Blocks Ranking Rgb & 3 & 6.00 & 4.00 & 6.00 & 1.00 & 11.00 & 2.00 & 10.00 & 6.00 & 31.00 & 3.00 & 63.00 & 8.00 \\
Blocks Ranking Size & 3 & 4.00 & 0.00 & 11.00 & 2.00 & 28.00 & 15.00 & 32.00 & 19.00 & 22.00 & 11.00 & 33.00 & 11.00 \\
Click Alarmclock & 1 & 78.00 & 49.00 & 95.00 & 65.00 & 91.00 & 69.00 & 83.00 & 55.00 & 100.00 & 67.00 & 22.00 & 62.00 \\
Click Bell & 1 & 76.00 & 65.00 & 97.00 & 95.00 & 99.00 & 89.00 & 93.00 & 76.00 & 91.00 & 95.00 & 6.00 & 85.00 \\
Dump Bin Bigbin & 1 & 7.00 & 1.00 & 87.00 & 61.00 & 80.00 & 77.00 & 95.00 & 75.00 & 85.00 & 74.00 & 97.00 & 30.00 \\
Grab Roller & 1 & 74.00 & 58.00 & 91.00 & 85.00 & 100.00 & 91.00 & 100.00 & 95.00 & 98.00 & 97.00 & 100.00 & 63.00 \\
Handover Block & 2 & 6.00 & 0.00 & 37.00 & 0.00 & 56.00 & 1.00 & 52.00 & 1.00 & 29.00 & 1.00 & 42.00 & 0.00 \\
Handover Mic & 2 & 63.00 & 1.00 & 70.00 & 1.00 & 53.00 & 1.00 & 43.00 & 1.00 & 62.00 & 0.00 & 37.00 & 2.00 \\
Hanging Mug & 2 & 1.00 & 0.00 & 6.00 & 0.00 & 5.00 & 1.00 & 15.00 & 2.00 & 18.00 & 9.00 & 19.00 & 1.00 \\
Lift Pot & 1 & 8.00 & 0.00 & 31.00 & 0.00 & 31.00 & 4.00 & 50.00 & 6.00 & 32.00 & 7.00 & 0.00 & 0.00 \\
Move Can Pot & 1 & 52.00 & 4.00 & 54.00 & 2.00 & 45.00 & 0.00 & 80.00 & 0.00 & 77.00 & 6.00 & 59.00 & 0.00 \\
Move Pillbottle Pad & 1 & 23.00 & 3.00 & 36.00 & 15.00 & 46.00 & 12.00 & 54.00 & 19.00 & 83.00 & 24.00 & 66.00 & 20.00 \\
Move Playingcard Away & 1 & 63.00 & 20.00 & 74.00 & 44.00 & 87.00 & 70.00 & 78.00 & 58.00 & 92.00 & 76.00 & 87.00 & 14.00 \\
Move Stapler Pad & 1 & 1.00 & 0.00 & 9.00 & 0.00 & 15.00 & 8.00 & 14.00 & 4.00 & 14.00 & 6.00 & 8.00 & 4.00 \\
Open Laptop & 1 & 66.00 & 18.00 & 81.00 & 24.00 & 88.00 & 51.00 & 88.00 & 51.00 & 92.00 & 60.00 & 14.00 & 3.00 \\
Open Microwave & 1 & 48.00 & 11.00 & 38.00 & 17.00 & 48.00 & 17.00 & 40.00 & 31.00 & 56.00 & 28.00 & 18.00 & 14.00 \\
Pick Diverse Bottles & 2 & 46.00 & 9.09 & 59.00 & 17.17 & 63.00 & 33.33 & 62.00 & 23.23 & 64.00 & 31.31 & 83.00 & 14.00 \\
Pick Dual Bottles & 2 & 52.00 & 34.00 & 75.00 & 39.00 & 71.00 & 54.00 & 63.00 & 40.00 & 86.00 & 45.00 & 86.00 & 20.00 \\
Place A2b Left & 1 & 29.00 & 6.00 & 53.00 & 6.00 & 32.00 & 8.00 & 57.00 & 27.00 & 62.00 & 17.00 & 64.00 & 12.00 \\
Place A2b Right & 1 & 18.00 & 4.00 & 43.00 & 11.00 & 41.00 & 6.00 & 51.00 & 32.00 & 54.00 & 18.00 & 59.00 & 6.00 \\
Place Bread Basket & 1 & 12.00 & 10.00 & 65.00 & 37.00 & 52.00 & 38.00 & 60.00 & 53.00 & 77.00 & 55.00 & 60.00 & 38.00 \\
Place Bread Skillet & 2 & 27.00 & 8.00 & 46.00 & 24.00 & 60.00 & 34.00 & 55.00 & 32.00 & 73.00 & 31.00 & 59.00 & 19.00 \\
Place Burger Fries & 2 & 67.00 & 62.00 & 82.00 & 70.00 & 92.00 & 86.00 & 82.00 & 84.00 & 81.00 & 73.00 & 66.00 & 45.00 \\
Place Can Basket & 2 & 41.00 & 3.00 & 39.00 & 5.00 & 55.00 & 11.00 & 54.00 & 3.00 & 65.00 & 21.00 & 53.00 & 7.00 \\
Place Cans Plasticbox & 2 & 21.00 & 19.00 & 53.00 & 28.00 & 59.00 & 50.00 & 58.00 & 50.00 & 84.00 & 68.00 & 28.00 & 27.00 \\
Place Container Plate & 1 & 81.00 & 39.00 & 87.00 & 43.00 & 83.00 & 42.00 & 88.00 & 58.00 & 88.00 & 57.00 & 90.00 & 58.00 \\
Place Dual Shoes & 2 & 19.00 & 5.00 & 35.00 & 16.00 & 68.00 & 35.00 & 60.00 & 37.00 & 63.00 & 19.00 & 46.00 & 3.00 \\
Place Empty Cup & 1 & 34.00 & 30.00 & 85.00 & 49.00 & 91.00 & 63.00 & 82.00 & 71.00 & 93.00 & 71.00 & 96.00 & 53.00 \\
Place Fan & 1 & 5.00 & 1.00 & 12.00 & 0.00 & 40.00 & 5.00 & 53.00 & 19.00 & 61.00 & 21.00 & 45.00 & 8.00 \\
Place Mouse Pad & 1 & 12.00 & 5.00 & 29.00 & 10.00 & 37.00 & 17.00 & 43.00 & 19.00 & 41.00 & 12.00 & 40.00 & 11.00 \\
Place Object Basket & 2 & 52.00 & 1.00 & 53.00 & 11.00 & 74.00 & 19.00 & 68.00 & 19.00 & 62.00 & 31.00 & 67.00 & 25.00 \\
Place Object Scale & 1 & 19.00 & 3.00 & 36.00 & 4.00 & 65.00 & 19.00 & 53.00 & 20.00 & 58.00 & 16.00 & 73.00 & 20.00 \\
Place Object Stand & 1 & 40.00 & 4.00 & 68.00 & 26.00 & 87.00 & 30.00 & 79.00 & 29.00 & 94.00 & 46.00 & 80.00 & 45.00 \\
Place Phone Stand & 1 & 15.00 & 6.00 & 34.00 & 14.00 & 45.00 & 18.00 & 62.00 & 29.00 & 50.00 & 22.00 & 54.00 & 7.00 \\
Place Shoe & 1 & 14.00 & 8.00 & 40.00 & 16.00 & 77.00 & 35.00 & 61.00 & 43.00 & 68.00 & 44.00 & 77.00 & 41.00 \\
Press Stapler & 1 & 71.00 & 39.00 & 85.00 & 62.00 & 84.00 & 64.00 & 78.00 & 56.00 & 82.00 & 58.00 & 61.00 & 72.00 \\
Put Bottles Dustbin & 3 & 17.00 & 9.00 & 40.00 & 14.00 & 70.00 & 56.00 & 64.00 & 40.00 & 59.00 & 50.00 & 63.00 & 10.00 \\
Put Object Cabinet & 2 & 28.00 & 4.00 & 50.00 & 5.00 & 49.00 & 6.00 & 57.00 & 12.00 & 61.00 & 12.00 & 34.00 & 7.00 \\
Rotate QRcode & 1 & 49.00 & 1.00 & 77.00 & 4.00 & 86.00 & 6.00 & 85.00 & 16.00 & 86.00 & 11.00 & 80.00 & 10.00 \\
Scan Object & 2 & 8.00 & 0.00 & 19.00 & 7.00 & 21.00 & 5.00 & 38.00 & 19.00 & 42.00 & 33.00 & 40.00 & 7.00 \\
Shake Bottle & 1 & 99.00 & 70.00 & 100.00 & 86.00 & 100.00 & 96.00 & 100.00 & 98.00 & 100.00 & 95.00 & 100.00 & 96.00 \\
Shake Bottle Horizontally & 1 & 100.00 & 76.00 & 100.00 & 92.00 & 100.00 & 94.00 & 100.00 & 98.00 & 100.00 & 94.00 & 100.00 & 94.00 \\
Stack Blocks Three & 3 & 5.00 & 2.00 & 10.00 & 1.00 & 48.00 & 11.00 & 46.00 & 19.00 & 31.00 & 7.00 & 54.00 & 14.00 \\
Stack Blocks Two & 2 & 38.00 & 6.00 & 55.00 & 13.00 & 93.00 & 39.00 & 94.00 & 37.00 & 62.00 & 21.00 & 85.00 & 33.00 \\
Stack Bowls Three & 3 & 52.00 & 23.00 & 74.00 & 42.00 & 74.00 & 44.00 & 66.00 & 42.00 & 68.00 & 44.00 & 70.00 & 24.00 \\
Stack Bowls Two & 2 & 84.00 & 51.00 & 92.00 & 75.00 & 90.00 & 59.00 & 88.00 & 72.00 & 91.00 & 63.00 & 93.00 & 52.00 \\
Stamp Seal & 1 & 12.00 & 8.00 & 39.00 & 17.00 & 57.00 & 25.00 & 38.00 & 26.00 & 61.00 & 20.00 & 48.00 & 15.00 \\
Turn Switch & 1 & 33.00 & 13.00 & 23.00 & 26.00 & 42.00 & 10.00 & 46.00 & 15.00 & 36.00 & 17.00 & 25.00 & 20.00 \\
\midrule
\textbf{Average} & -- & \textbf{37.86} & \textbf{16.80} & \textbf{53.92} & \textbf{26.74} & \textbf{63.28} & \textbf{34.23} & \textbf{64.16} & \textbf{37.16} & \textbf{67.30} & \textbf{37.77} & \textbf{57.00} & \textbf{25.68} \\
\bottomrule
\end{tabular}%
}
\end{table*}

\end{document}